\documentclass[lettersize,journal]{IEEEtran}
\usepackage{amsmath,amsfonts}
\usepackage{bm}
\usepackage{algorithm}
\usepackage{array}
\usepackage[caption=false,font=normalsize,labelfont=sf,textfont=sf]{subfig}
\usepackage{textcomp}
\usepackage{stfloats}
\usepackage{url}
\usepackage{verbatim}
\usepackage{graphicx}
\usepackage{cite}
\usepackage{color}
\usepackage{booktabs}
\usepackage{algorithm, algpseudocode}
\usepackage{hyperref}
\algrenewcommand\alglinenumber[1]{\tiny #1:}

\newcommand\blfootnote[1]{%
  \begingroup
  \renewcommand\thefootnote{}\footnote{#1}%
  \addtocounter{footnote}{-1}%
  \endgroup
}

\definecolor{darkgreen}{rgb}{0,0.5,0.7}
\definecolor{lightgreen}{rgb}{0.4,0.8,0.4}
\definecolor{lightorange}{rgb}{0.9,0.6,0.1}
\definecolor{darkpurple}{rgb}{0.5,0,0.7}
\definecolor{darkblue}{rgb}{0.5,0,0.7}
\definecolor{modify}{rgb}{1, 0.5, 0}
\definecolor{StaticNet}{rgb}{0.62,0.478,0.259}
\definecolor{DynamicNet}{rgb}{0.216,0.537,0.620}
\definecolor{RayRefinement}{rgb}{0.494,0.392,0.620}
\definecolor{Fthetal}{rgb}{0.573,0.224,0.196}
\definecolor{Sgtq}{rgb}{0.553,0.486,0.357}
\definecolor{St}{rgb}{0.749,0.565,0}
\definecolor{lightorange}{rgb}{0.9,0.6,0.1}

\newcommand{\jm}[1]  {{#1}}
\newcommand{\quan}[1] {{#1}}

\algnewcommand\algorithmicforeach{\textbf{for each}}
\algdef{S}[FOR]{ForEach}[1]{\algorithmicforeach\ #1\ \algorithmicdo}

\begin{document}

\title{MoBluRF: Motion Deblurring Neural Radiance Fields for Blurry Monocular Video}

\author{Minh-Quan~Viet~Bui\textsuperscript{1*}, Jongmin~Park\textsuperscript{1*}, Jihyong~Oh\textsuperscript{2\textdagger}, and Munchurl~Kim\textsuperscript{1\textdagger},~\IEEEmembership{Senior Member,~IEEE,}

\textsuperscript{1} School of Electrical Engineering, Korea Advanced Institute of Science and Technology \\
\textsuperscript{2} Department of Imaging Science, GSAIM, Chung-Ang University
}

\markboth{Journal of \LaTeX\ Class Files,~Vol.~14, No.~8, August~2021}%
{Shell \MakeLowercase{\textit{et al.}}: A Sample Article Using IEEEtran.cls for IEEE Journals}

\IEEEpubid{0000--0000/00\$00.00~\copyright~2021 IEEE}

\maketitle

\begin{abstract}
Neural Radiance Fields (NeRF), initially developed for static scenes, have inspired many video novel view synthesis techniques. However, the challenge for video view synthesis arises from motion blur, a consequence of object or camera movements during exposure, which hinders the precise synthesis of sharp spatio-temporal views. In response, we propose a novel motion deblurring NeRF framework for blurry monocular video, called MoBluRF, consisting of a Base Ray Initialization (BRI) stage and a Motion Decomposition-based Deblurring (MDD) stage. In the BRI stage, we coarsely reconstruct dynamic 3D scenes and jointly initialize the base rays which are further used to predict latent sharp rays, using the inaccurate camera pose information from the given blurry frames. In the MDD stage, we introduce a novel Incremental Latent Sharp-rays Prediction (ILSP) approach for the blurry monocular video frames by decomposing the latent sharp rays into global camera motion and local object motion components. We further propose two loss functions for effective geometry regularization and decomposition of static and dynamic scene components without any mask supervision. Experiments show that MoBluRF outperforms qualitatively and quantitatively the recent state-of-the-art methods with large margins.
\end{abstract}


\vspace{-5mm}
\section{Introduction}
\IEEEPARstart{F}{ree} \quan{viewpoint rendering for spatio-temporal novel view synthesis (NVS) has gained significant attention due to its broad applicability in VR, telepresence, and video production. In particular, video NVS enables the generation of photorealistic frames from arbitrary camera viewpoints and time instances, allowing for compelling reconstructions of dynamic scenes. Building upon the introduction of Neural Radiance Fields (NeRF)~\cite{mildenhall2020nerf} for static scene representation, a wide variety of methods have been actively explored to extend NVS to the temporal domain using monocular or multi-view setups \cite{pumarola2021d, li2022neural, wang2022fourier, weng2022humannerf, zhang2003spacetime, zitnick2004high, oswald2014generalized, collet2015high, broxton2020immersive, li2021neural, gao2021dynamic, tretschk2021non, li2023dynibar, attal2023hyperreel, park2023temporal, shao2023tensor4d, fridovich2023k}. Despite these advances, the existing methods \cite{pumarola2021d, li2022neural, wang2022fourier, weng2022humannerf, zhang2003spacetime, zitnick2004high, oswald2014generalized, collet2015high, broxton2020immersive, li2021neural, gao2021dynamic, tretschk2021non, li2023dynibar, attal2023hyperreel, park2023temporal, shao2023tensor4d, fridovich2023k} remain highly sensitive to the quality of training videos \cite{wang2023bad, lee2023dp, ma2022deblur, sun2024dyblurf}. In particular, motion blur—caused by fast object motion \cite{pan2016blind, zhang2020deblurring} or camera shakes \cite{bahat2017non, zhang2018adversarial}—commonly occurs due to light accumulation during video exposure \cite{telleen2007synthetic, gupta2010single, harmeling2010space, oh2022demfi}, and is prevalent in real-world videos such as mobile phone footage, sports broadcasts, or drone captures. Consequently, generating \textit{sharp} novel spatio-temporal views from monocular videos remains a challenging task, especially in the presence of \textit{blurriness} introduced during the capture process. This work specifically targets this challenge by aiming to render sharp NVS from blurry monocular videos.}
\blfootnote{*Co-first authors (equal contribution)}
\blfootnote{\textdagger Co-corresponding authors}

\quan{Reconstructing sharp dynamic NVS results from blurry monocular videos poses several key challenges: First, the deblurring process must preserve 3D-consistent spatial structures across frames. Using 2D deblurring methods~\cite{zhu2023exploring, pan2023deep, pan2020cascaded, wang2022MMP, zhou2019spatio, deng2021multi, liang2022rvrt, zhang2022spatio, li2023simple, zhong2021towards, zamir2022restormer, bahat2017non, chakrabarti2016neural, cho2009fast, schuler2015learning, shan2008high, xu2010two, sun2015learning, fang2023self} as preprocessing for blurry input videos prior to dynamic NVS optimization may lead to geometric inconsistencies that cannot be corrected in the view synthesis stage; Second, the deblurring process should account for both global camera motion and local object motion over time, especially when predicting multiple rays per pixel to model motion blur—commonly referred to as latent sharp rays. Most existing deblurring-based NVS methods~\cite{ma2022deblur, wang2023bad, lee2023dp, lee2023exblurf}, however, have been developed for static multi-view settings and fail to incorporate such temporal dynamics, making them ineffective when applied to blurry monocular videos; Lastly, blurry videos often result in inaccurate camera pose estimations, which can destabilize the joint optimization of dynamic radiance fields and latent sharp ray prediction.}
\IEEEpubidadjcol

To address the challenges faced by the existing NVS methods with blurry videos, we propose MoBluRF, a novel and effective two-stage motion deblurring method for neural radiance fields. To enhance the stability of spatio-temporal sharp reconstruction from motion-blurry videos, we introduce a Base Ray Initialization (BRI) stage coupled with a novel interleaved optimization method. This BRI stage significantly enhances the robustness of the learned radiance fields, enabling the precise decomposition of target scenes into static and dynamic regions. It also produces optimized camera rays that greatly improve deblurring in subsequent stages. Moreover, we introduce a Motion Decomposition-based Deblurring (MDD) stage to adaptively handle blurriness resulting from a mixture of complex camera and object motions by simulating the dynamic blur process. To achieve this, our model incrementally decomposes global camera movements and local object motions. Additionally, we introduce two novel loss functions: $\mathcal{L}_{sm}$, which eliminates the dependency of motion mask preprocessing~\cite{liu2023robust} in an unsupervised manner, and $\mathcal{L}_{lg}$, which enhances the geometric accuracy of dynamic objects.
\jm{Our MoBluRF \textit{significantly} outperforms the \textit{most recent} dynamic deblurring NeRF method, DyBluRF~\cite{sun2024dyblurf}, and other SOTA methods such as static deblurring NeRFs~\cite{wang2023bad,lee2023dp}, dynamic NVS methods \cite{gao2022monocular, wu20234d, park2021hypernerf, fang2022fast, cao2023hexplane}, the cascades of 2D deblurring methods~\cite{li2023simple, zamir2022restormer, zhong2021towards} and dynamic NVS methods~\cite{park2021hypernerf, gao2022monocular, wu20234d, liu2023robust, li2021neural} (Table \ref{table:quantitative_comparison}, Table \ref{table:quantitative_comparison_stereo}).} Our contributions are summarized as follows:

\begin{figure*}[t]
  \centering 
  \includegraphics[scale=0.3]{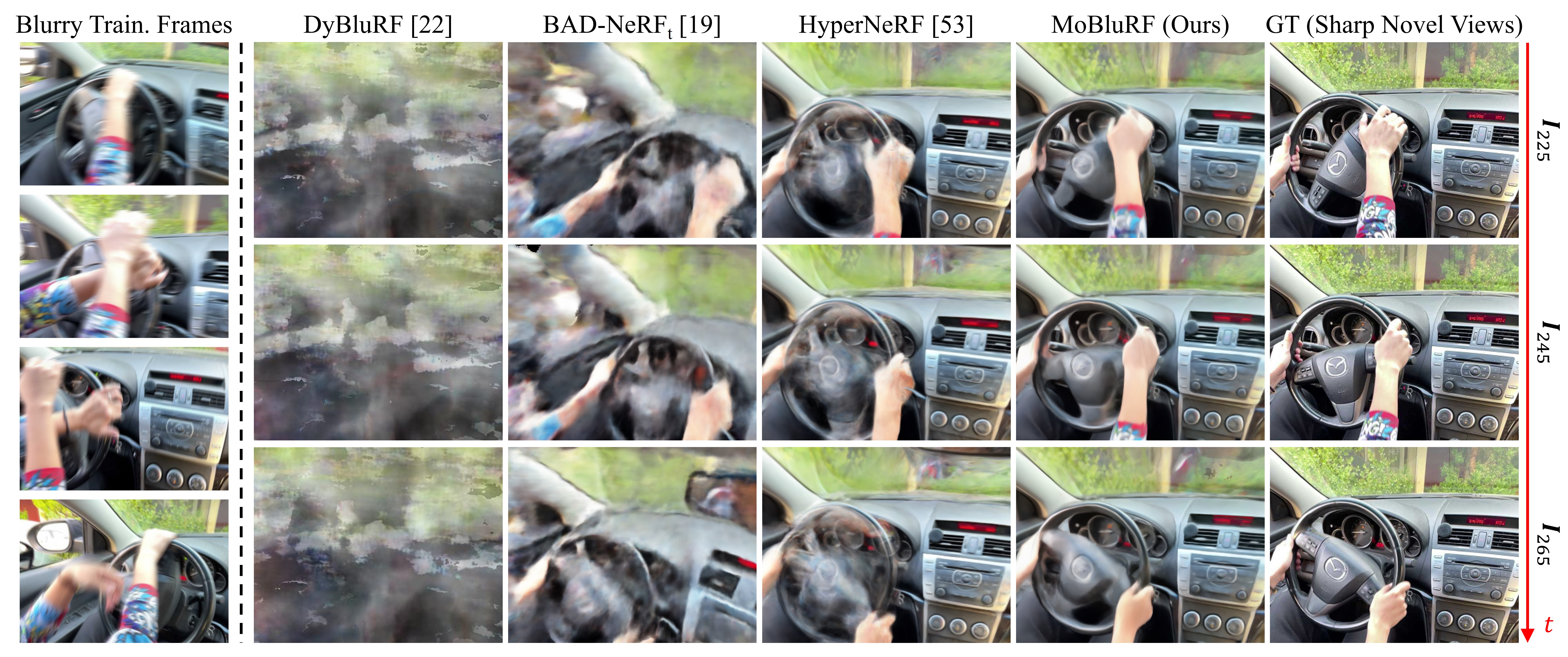}
  \vspace{-3mm}
  \caption{Motion deblurring novel view synthesis results. We propose a novel motion deblurring NeRF for \textit{blurry} monocular videos, called MoBluRF, which significantly outperforms previous SOTA NeRF methods, trained on the newly synthesized Blurry iPhone dataset.}
  \vspace{-5mm}
  \label{fig:figure_page1}
\end{figure*}

\begin{itemize}
 \setlength\itemsep{0.1cm}
  \item \quan{We propose a novel two-stage motion deblurring NeRF framework, called MoBluRF, to effectively render sharp novel spatio-temporal views from blurry monocular videos.} 
  \item \quan{The first stage of our framework is designed to coarsely reconstruct dynamic radiance fields from imprecise initial camera poses to mitigate unreliable geometry initialization from blurry monocular videos.}
  \item \quan{The second stage is designed to predict latent sharp rays incrementally to model the complex motion blur considering both global camera and local object movement.}
  \item \quan{We propose two simple yet effective loss functions for optimizing sharp radiance fields without any mask supervision, and for reconstructing robust geometry.}
  \item \jm{For experiments, we synthesize a new blurry version of iPhone dataset \cite{gao2022monocular} which is used to train MoBluRF and other methods under fair comparison. The experimental results on various datasets, including our new dataset, show that MoBluRF outperforms SOTA methods, both qualitatively (Fig. \ref{fig:figure_page1}, Fig. \ref{fig:figure_qualitative}, Fig. \ref{fig:figure_qualitative_dyblurf}, Fig. \ref{fig:figure_qualitative_davis}) and quantitatively (Table \ref{table:quantitative_comparison}, Table \ref{table:quantitative_comparison_stereo}) in motion deblurring for novel view synthesis. Our MoBluRF is robust against various degrees of blurriness (Fig. \ref{fig:figure_qualitative_robust}, Fig. \ref{fig:figure_robust}, Fig. \ref{fig:figure_robust_extended}, Table \ref{table:quantitative_robust}).}
\end{itemize}

\section{Related Work}
\subsection{NVS for Dynamic Scenes}
Recent methods for video view synthesis have expanded upon the static NeRF framework \cite{mildenhall2020nerf}. They represent dynamic NeRFs by incorporating scene flow-based frameworks \cite{gao2021dynamic, li2021neural, li2023dynibar} or canonical fields \cite{park2021nerfies, park2021hypernerf, pumarola2021d, tretschk2021non, weng_humannerf_2022_cvpr, jiang2022neuman, yang2022banmo, athar2022rignerf, fang2022fast, song2023nerfplayer, liu2023robust} to model non-rigid deformable transformations or 4D spatio-temporal radiance fields \cite{xian2021space, li2021neural, gao2021dynamic, du2021neural, li2022neural, van2022revealing, cao2023hexplane, fridovich2023k, shao2023tensor4d, attal2023hyperreel}. The methods such as NSFF \cite{li2021neural}, DynamicNeRF \cite{gao2021dynamic}, and DynIBaR \cite{li2023dynibar} typically combine two types of NeRFs: time-invariant and time-variant, to generate novel spatio-temporal views for monocular videos. However, they rely heavily on pretrained motion mask extraction for moving objects and various regularization losses for 3D scene flows, which makes them less effective in deblurring video view synthesis. T-NeRF~\cite{gao2022monocular}, HyperNeRF~\cite{park2021hypernerf}, TiNeuVox~\cite{fang2022fast} initially learn deformation or offset fields that transform a ray in an observation space to a bent ray in a canonical space. 
\jm{Building on the success of 3DGS~\cite{kerbl20233d} in achieving faster and higher-fidelity novel view synthesis, several methods~\cite{wu20234d, duan20244d, stearns2024dynamic, chu2024dreamscene4d, wang2024shape, lei2024mosca} that model time-varying Gaussian attributes have been proposed for video view synthesis. However, none of the above existing SOTA methods can be readily applied to deblurring-based novel view synthesis from the given blurry frames due to the lack of an effective deblurring strategy.}

\vspace{-0.2cm}
\subsection{Deblurring NVS}
To produce visually appealing frames with consistent 3D geometry from blurry multi-view \textit{static} images, several deblurring NVS methods have emerged. DeblurNeRF \cite{ma2022deblur} employs an end-to-end volume rendering framework \cite{drebin1988volume} to estimate spatial blur kernels at the pixel level and the latent sharp radiance fields; BAD-NeRF \cite{wang2023bad} jointly predicts the virtual camera trajectories during the image exposure capture time. DP-NeRF \cite{lee2023dp} introduces a rigid blurring kernel to maintain 3D consistency by leveraging physical constraints. ExBluRF \cite{lee2023exblurf} adopts an MLP-based framework to reduce the dimensionality of 6-DOF camera poses and utilizes a voxel-based radiance field \cite{chen2022tensorf,fridovich2022plenoxels}. \jm{More recently, a 3DGS~\cite{kerbl20233d}-based deblurring NVS~\cite{peng2025bags} method proposes a blur agnostic architecture that utilizes per-pixel convolution kernels. Nonetheless, none of the above methods \textit{are applicable to non-rigid video view synthesis} due to the lack of motion-aware deblurring in the temporal dimension.}

\jm{Very recently, to address the blurriness caused by camera and object motion in video view synthesis, several methods~\cite{sun2024dyblurf, luthra2024deblur, luo2024dynamic} have been proposed.} The most representative dynamic deblurring NeRF method, DyBluRF~\cite{sun2024dyblurf}, models the blurry RGB color by averaging sharp volume-rendered colors at each discrete timestamp by considering the camera and object motions. To account for the moving objects' global trajectory, the method utilizes the Discrete Cosine Transform (DCT)~\cite{valmadre2012general}.
DyBluRF~\cite{sun2024dyblurf} is significantly different from our MobluRF in several aspects: (i) Similar to the static deblurring NeRFs~\cite{wang2023bad, lee2023dp, lee2023exblurf}, the DyBluRF~\cite{sun2024dyblurf} predicts camera and motion trajectories during exposure time directly from inaccurate camera poses. Additionally, it simultaneously optimizes these trajectories in conjunction with the dynamic radiance fields. We observe that these strategies lead to unstable optimization (Table \ref{table:ablation_study}) for long-length blurry videos with a significant mixture of camera and object motions. In contrast, we propose a novel Base Ray Initialization (BRI) stage (Sec. \ref{sect:base_ray_initial_stage}) with an interleaved optimization scheme, which significantly improves the reconstruction results with sharp radiance fields; (ii) The DyBluRF~\cite{sun2024dyblurf} relies heavily on supplementary priors such as motion masks~\cite{liu2023robust}, depths~\cite{ranftl2020towards}, and optical flows~\cite{teed2020raft}, which are challenging to estimate precisely from \textit{motion-blurry} videos. In contrast, we introduce two novel loss functions, $\mathcal{L}_{sm}$, for effectively decomposing static and dynamic radiance fields in an \textit{unsupervised} manner, and $\mathcal{L}_{lg}$ for reconstructing the robust dynamic object geometries from scale- and shift-ambiguous depth maps.

\section{Proposed Method: MoBluRF}
\subsection{Design Considerations}
\label{sect:design_consideration}
\begin{figure*}[t]
\centering
\includegraphics[scale=0.46]{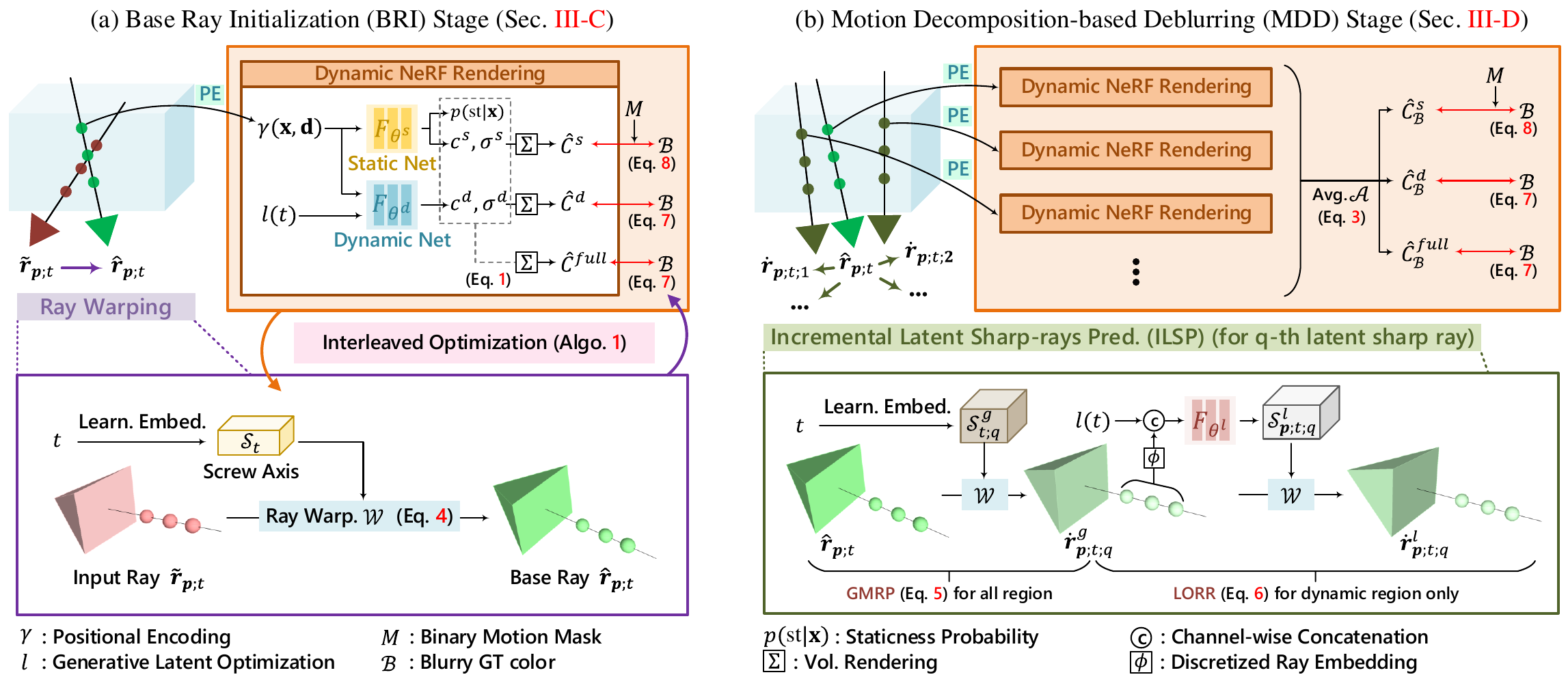}
\vspace{-0.2cm}
\caption{Overview of our MoBluRF framework. To effectively optimize the sharp radiance field with the imprecise camera poses from blurry video frames, we design our MoBluRF consisting of two main procedures (Algo. \ref{alg:overall_process}) of (a) \textit{Base Ray Initialization (BRI) Stage} (Sec. \ref{sect:base_ray_initial_stage} and Algo. \ref{alg:bri_stage}) and (b) \textit{Motion Decomposition-based Deblurring (MDD) Stage} (Sec. \ref{sect:deblur_module}).}
\label{fig:MoBluRF}
\end{figure*}

Our MoBluRF is designed to represent \textit{sharp} dynamic neural radiance fields from \textit{blurry} monocular videos, which consists of two main stages: Base Ray Initialization (BRI) and Motion Decomposition-based Deblurring (MDD), as shown in Fig. \ref{fig:MoBluRF}. We provide detailed descriptions of our MoBluRF process, including pseudo-codes for three algorithms (Algo. \ref{alg:bri_stage}, Algo. \ref{alg:overall_process} and Algo. \ref{alg:inference}), in the following subsections.

To reconstruct sharp dynamic radiance fields from blurry monocular videos, an accurate prediction of latent sharp rays emitted from camera and motion trajectories during exposure time is essential~\cite{wang2023bad, lee2023dp, ma2022deblur, sun2024dyblurf}. This process aims to render the values of blurry pixels in accordance with the physical process~\cite{nah2017deep, Nah_2019_CVPR_Workshops_REDS}. Typically, latent sharp rays are estimated by calculating the residual transformation of a ray that we refer to as a \textit{base ray}~\cite{lee2023dp}. However, if we directly utilize the input rays which are derived from the camera poses obtained from blurry videos as the base rays, incorrect latent sharp rays \cite{wang2023bad} may be predicted. Moreover, in video view synthesis, jointly optimizing dynamic radiance fields and predicting latent sharp rays from imprecise base rays can lead to suboptimal solutions. To overcome these issues, distinguished from previous deblurring NeRFs \cite{ma2022deblur, wang2023bad, lee2023dp, sun2024dyblurf} that directly predict latent sharp rays from imprecisely estimated camera poses, we \textit{newly} propose the BRI stage (Sec. \ref{sect:base_ray_initial_stage}) to provide a stronger foundation for the accurate prediction of latent sharp rays. The BRI stage coarsely reconstructs robust dynamic radiance fields and refines the initialization of base rays from imprecisely obtained camera rays.

On the other hand, the existing SOTA static deblurring NeRFs \cite{ma2022deblur, wang2023bad, lee2023dp, lee2023exblurf} only consider \textit{multi-view} static images, so facing with the difficulties in capturing temporal information for blurry \textit{monocular} videos due to the absence of a motion-aware deblurring module. Other SOTA methods \cite{park2021hypernerf,fang2022fast,cao2023hexplane,wu20234d} for monocular video view synthesis are not capable of handling input blurry frames due to lack of deblurring components. To overcome these limitations, we \textit{newly} introduce the MDD stage (Sec. \ref{sect:deblur_module}), with a novel Incremental Latent Sharp-rays Prediction (ILSP) method which can effectively synthesize the physical blur process considering global and local motions in a progressive manner along temporal axis.

\vspace{-0.2cm}
\subsection{Preliminaries}
\label{sect:preliminary}
\noindent \textbf{Dynamic Neural Radiance Fields}
We extend the static NeRF \cite{mildenhall2020nerf} to our MoBluRF for the monocular video which consists of one frame per time $t$. Our MoBluRF learns to represent the continuous radiance of a video scene using neural networks, taking into account a set of $N_f$ frames from the monocular video, denoted as $\{\bm{\mathcal{I}}_{t}\}_{t=1}^{N_f}$, and the corresponding camera poses $\{\bm{\mathcal{P}}_{t}\}_{t=1}^{N_f}$.
Following \cite{gao2021dynamic, li2021neural, li2023dynibar, liu2023robust}, we decompose our radiance representation into Static Net $F_{\theta^s}$ and Dynamic Net $F_{\theta^d}$. Given a 3D position $\mathbf{x} = (x,y,z)$ and a viewing direction $\mathbf{d}$, $F_{\theta^s}: \gamma(\mathbf{x}, \mathbf{d}) \rightarrow (\bm{c}^s, \sigma^s, p(\text{st}|\mathbf{x}))$ estimates a color $\bm{c}^s$, a volume density $\sigma^s$ of static scene components (static regions) and a staticness probability $p(\text{st}|\mathbf{x})$ that $\mathbf{x}$ is static from the spatial positional encoded inputs $\gamma(\mathbf{x}, \mathbf{d})$. On the other hand, $F_{\theta^d}: (\gamma(\mathbf{x}, \mathbf{d}), l(t)) \rightarrow (\bm{c}^d, \sigma^d)$ maps a time-varying embedding to a color $\bm{c}^d$ and a volume density $\sigma^d$ of dynamic scene components (dynamic regions) where $\gamma$ is the positional encoding~\cite{mildenhall2020nerf} and $l$ is Generative Latent Optimization (GLO) \cite{glo}. 
Let $\bm{r}_{\bm{p};t}(\kappa) = \mathbf{o}_t + \kappa\mathbf{d}_{\bm{p};t}$ be the cast ray from a camera origin $\mathbf{o}_t$ through a given pixel $\bm{p}$ of the image plane at time $t$ where $\kappa$ and $\mathbf{d}_{\bm{p};t}$ denote a sampling ray distance and a viewing direction through pixel $\bm{p}$ at time $t$, respectively. We separately estimate the rendered colors $\hat{\bm{C}}^s(\bm{r}_{\bm{p};t})$ of the static scene components and $\hat{\bm{C}}^d(\bm{r}_{\bm{p};t})$ of the dynamic scene components via continuous volume rendering \cite{drebin1988volume} by approximately computing the integral on $N$ piecewise constant segments $\{[\rho_{n}, \rho_{n+1}]\}^N_{n=1}$ along the ray $\bm{r}_{\bm{p};t}$ as $\hat{\bm{C}}^s(\bm{r}_{\bm{p};t}) = \sum_{n=1}^{N} \mathcal{T}^s_n  \alpha^s_n  \bm{c}^s_n$ and $\hat{\bm{C}}^d(\bm{r}_{\bm{p};t}) = \sum_{n=1}^{N} \mathcal{T}^d_n  \alpha^d_n  \bm{c}^d_n$
where $\mathcal{T}_n$ is the accumulated transmittance and $\alpha_n$ is the alpha-compositing weight. Here, $\alpha_n = 1-\text{exp}(-\sigma_n\delta_n)$ and $\mathcal{T}_n = \prod_{k=1}^{n-1} 1 -\alpha_k$
where $\delta_n = \rho_{n+1} - \rho_n$ is the length of the segment. To predict the full rendered color $\hat{\bm{C}}^{full}(\bm{r}_{\bm{p};t})$ of pixel $\bm{p}$ with camera pose $\bm{\mathcal{P}}_{t}$, MoBluRF combines the outputs of $F_{\theta^s}$ and $F_{\theta^d}$ via the probabilistic volume rendering as:
\begin{equation}
\label{eq:full_color_rendering}
\begin{split}
        \hat{\bm{C}}^{full}(\bm{r}_{\bm{p};t}) = & \textstyle\sum_{n=1}^{N}  \mathcal{T}^{full}_n p(\text{st}|\mathbf{x}_n) \alpha^s_n  \bm{c}^s_n  \\
        + & \textstyle\sum_{n=1}^{N}  \mathcal{T}^{full}_n (1-p(\text{st}|\mathbf{x}_n))  \alpha^d_n  \bm{c}^d_n,
\end{split}
\end{equation}
where $\mathbf{x}_n$ is the $n^{\text{th}}$ 3D point and the full accumulated transmittance $\mathcal{T}^{full}_n$ is calculated as $\prod_{k=1}^{n-1} (1 -p(\text{st}|\mathbf{x}_k)\alpha^s_k)(1 -(1-p(\text{st}|\mathbf{x}_k))\alpha^d_k)$.

\noindent \textbf{Binary Motion Mask Prediction.}
Learning the motion decomposition has been widely adopted in previous works \cite{gao2021dynamic, li2023dynibar, li2022neural} to stabilize the reconstruction of static scene components in the dynamic NeRFs. In our MoBluRF, we also predict the 2D binary motion mask for the BRI stage and the MDD stage optimizations. The binary motion mask $\bm{M}(\bm{r}_{\bm{p};t})$ can be obtained by thresholding the dynamicness probability $p(\text{dy}|\bm{r}_{\bm{p};t}) > 0.5$, 
where $p(\text{dy}|\bm{r}_{\bm{p};t})$ is the accumulated dynamicness probability of all sampling points $\mathbf{x}_n$ along $\bm{r}_{\bm{p};t}$ as:
\vspace{-0.2cm}
\begin{equation}
\begin{split}
       p(\text{dy}|\bm{r}_{\bm{p};t}) \hspace{-0.1cm} &= \hspace{-0.1cm} \textstyle\sum_{\mathbf{x}_n }p(\mathbf{x}_n|\bm{r}_{\bm{p};t})p(\text{dy}|\mathbf{x}_n,\bm{r}_{\bm{p};t}) \hspace{-0.1cm} \\ &= \hspace{-0.1cm} \textstyle\sum_{\mathbf{x}_n}p(\mathbf{x}_n|\bm{r}_{\bm{p};t})(1-p(\text{st}|\mathbf{x}_n,\bm{r}_{\bm{p};t})),
\end{split}
\label{eq:pdy}
\end{equation}
where we assume that $p(\text{st}|\mathbf{x}_n, \bm{r}_{\bm{p};t}) = p(\text{st}|\mathbf{x}_n)$, as the staticness probability of the 3D sampling point $\mathbf{x}_n$ should remain consistent regardless of the cast ray and the time index. Thus, using $p(\text{st}|\mathbf{x})$ from $F_{\theta^s}$ suffices to estimate the binary motion mask. $p(\mathbf{x}_n|\bm{r}_{\bm{p};t})$ is computed from accumulated transmittance $\mathcal{T}_n$ and alpha-compositing weight $\alpha_n$.

\noindent \textbf{Deblurring Neural Radiance Fields.}
To solve the blur problem for the reconstruction of sharp radiance fields, we simulate the physical blur process similar to the existing deblurring NeRF methods~\cite{ma2022deblur, lee2023dp, wang2023bad, sun2024dyblurf}. The physical blur process which generates a blurry color $\bm{\mathcal{B}}_{\bm{p};t}$ of pixel $\bm{p}$ at time $t$ by applying an unknown motion blur kernel $k_{\bm{p};t}$ along the spatio-temporal direction to the set of sharp pixel colors $\bm{\mathcal{I}}_{\bm{p};t}$ is formulated as $\bm{\mathcal{B}}_{\bm{p};t} = k_{\bm{p};t} \ast \bm{\mathcal{I}}_{\bm{p};t}$ where $\ast$ indicates the spatio-temporal convolution operation \cite{Nah_2019_CVPR_Workshops_REDS, nah2017deep,shen2020blurry,oh2022demfi,wang2023bad}.
To effectively train our MoBluRF for the given blurry monocular video frames $\{\bm{\mathcal{B}}_{t}\}_{t=1}^{N_f}$ with the resultant inaccurate camera poses $\{\tilde{\bm{\mathcal{P}}}_{t}\}_{t=1}^{N_f}$,
we model the blur process for monocular dynamic radiance fields by predicting the set of latent sharp rays $\{\dot{\bm{r}}_{\bm{p};t;q}\}_{q=1}^{N_b}$ casting based on the base ray $\bm{r}_{\bm{p};t}$.
Then, we average the corresponding volume rendered pixel colors to generate a blurry pixel color where $q$ is the index and $N_b$ is the number of latent sharp rays, respectively. We denote this motion blur process as:
\begin{equation}
\begin{split}
    \hat{\bm{C}}_\mathcal{B}(\bm{r}_{\bm{p};t}) &= \mathcal{A}(\hat{\bm{C}}(\bm{r}_{\bm{p};t}), \{\hat{\bm{C}}(\dot{\bm{r}}_{\bm{p};t;q})\}_{q=1}^{N_b}) \\
    &= \frac{1}{N_b+1} \Bigl(\hat{\bm{C}}(\bm{r}_{\bm{p};t}) + \textstyle\sum_{q=1}^{N_b} \hat{\bm{C}}(\dot{\bm{r}}_{\bm{p};t;q})\Bigr),
\end{split}
\label{eq:avg_color}
\end{equation}
where $\hat{\bm{C}}_\mathcal{B}(\bm{r}_{\bm{p};t})$ is a blurry rendered color of the base ray $\bm{r}_{\bm{p};t}$ and $\mathcal{A}(\cdot ,\cdot)$ is an average function of the sharp rendered color $\hat{\bm{C}}(\bm{r}_{\bm{p};t})$ of $\bm{r}_{\bm{p};t}$ and the set of rendered colors $\{\hat{\bm{C}}(\dot{\bm{r}}_{\bm{p};t;q})\}_{q=1}^{N_b}$ of the corresponding latent sharp rays $\{\dot{\bm{r}}_{\bm{p};t;q}\}_{q=1}^{N_b}$.

\subsection{Base Ray Initialization (BRI) Stage}
\label{sect:base_ray_initial_stage}
Unlike the existing deblurring NeRF methods \cite{wang2023bad, lee2023dp, sun2024dyblurf} that directly estimate latent sharp rays during the exposure time, our proposed BRI stage first initializes the base rays to predict accurate latent sharp rays by warping them from the input rays. This initialization is jointly trained with $F_{\theta^s}$ and $F_{\theta^d}$ via our proposed interleaved optimization strategy, as shown in Fig. \ref{fig:MoBluRF}-(a). Then, our MoBluRF effectively predicts the latent sharp rays from these base rays in the following Motion Decomposition-based Deblurring (MDD) stage which is described in Sec. \ref{sect:deblur_module}. We now detail our ray warping and interleaved optimization in the BRI stage.

\noindent \textbf{Ray Warping.}
\label{sect:ray_refinement}
Let $\tilde{\bm{r}}_{\bm{p};t}$ represent an input ray emitted from pixel $\bm{p}$ with a camera pose $\tilde{\bm{\mathcal{P}}}_t$ from the blurry input video frame at time $t$. We initialize the base ray $\hat{\bm{r}}_{\bm{p};t}$ from the input ray $\tilde{\bm{r}}_{\bm{p};t}$ using a ray warping $\mathcal{W}$ as:
\begin{equation}
        \hat{\bm{r}}_{\bm{p};t} = \mathcal{W}(\tilde{\bm{r}}_{\bm{p};t}, \mathcal{S}_t) = e^{[\bm{\omega}_t]_\times}\tilde{\bm{r}}_{\bm{p};t} + \bm{G}_t\bm{v}_t,
\label{eq:bundle_adjust_ray}
\end{equation}
where $\mathcal{S}_t = (\bm{\omega}_t; \bm{v}_t) \in \mathbb{R}^6$ is a learnable screw axis with the corresponding rotation encoding $\bm{\omega}_t \in \mathbb{R}^3$ and translation encoding $\bm{v}_t \in \mathbb{R}^3$ at time $t$ \cite{park2021nerfies}.
The initialized base ray $\hat{\bm{r}}_{\bm{p};t}$ is specifically used as the term $\bm{r}_{\bm{p};t}$ in Eq. \ref{eq:avg_color}.
Similar to the existing methods \cite{park2021nerfies, lee2023dp}, we compute the residual rotation $e^{[\bm{\omega}_t]_\times} =  \textbf{I} + \frac{\sin{\theta}}{\theta}[\bm{\omega}_t]_\times + \frac{1-\cos{\theta}}{\theta^2}[\bm{\omega}_t]^2_\times$ and translation matrix $\bm{G}_t = \textbf{I} + \frac{1-\cos{\theta}}{\theta^2}[\bm{\omega}_t]_\times + \frac{\theta - \sin{\theta}}{\theta^3}[\bm{\omega}_t]^2_\times$, where $\textbf{I}$ is the $3\times3$ identity matrix, $[\mathbf{w}]_\times$ is the cross-product matrix of a $3\times1$ vector $\mathbf{w}$ and $\theta = \lVert \bm{\omega}_t \rVert$ is the rotation angle at time $t$. We model $\mathcal{S}_t$ as a learnable embedding of time $t$, as in Fig. \ref{fig:MoBluRF}-(a).

\noindent \textbf{Interleaved Optimization.}
\label{sect:interleaved_optim}
Jointly optimizing $F_{\theta^d}$ and ray warping is highly ill-posed and can lead to undesirable local minima (Variant (g) in Table \ref{table:ablation_study}). This issue arises because $F_{\theta^d}$ is prone to misinterpreting geometric inconsistencies, stemming from inaccurate base rays, as deformations of dynamic objects across time instances. This misinterpretation leads to unnecessary deformations predicted by $F_{\theta^d}$ and additional errors in predicting latent sharp rays.
To address this, we propose a novel interleaved optimization strategy that alternates between optimizing the screw axis $\mathcal{S}_t$ for ray warping and $F_{\theta^d}$, as shown in Fig. \ref{fig:MoBluRF}-(a) and Algo. \ref{alg:bri_stage}. More specifically, for the \textit{even} iteration indices, we jointly update $F_{\theta^s}$ and $\mathcal{S}_t$ using the static geometry cues extracted by our 2D binary motion mask $\bm{M}(\hat{\bm{r}}_{\bm{p};t})$. For the \textit{odd} iteration indices, we update both $F_{\theta^s}$ and $F_{\theta^d}$, and keep $\mathcal{S}_t$ unchanged. This approach enhances the training stability by simultaneously advancing the learning of 3D dynamic reconstruction and initializing the base rays for deblurring in the MDD stage.

\begin{algorithm}
\footnotesize
\caption{Base Ray Initialization (BRI) Stage}
\begin{algorithmic}[1]
\Procedure{BRI}{\textcolor{StaticNet}{$F_{\theta^s}$}, \textcolor{DynamicNet}{$F_{\theta^d}$}, \textcolor{St}{$\mathcal{S}_t$}}
\For{it $=0$ to $2\times10^5$}
    \State Sample random rays $\tilde{\bm{r}}_{\bm{p};t}$
    \State Compute base ray $\hat{\bm{r}}_{\bm{p};t} \leftarrow \mathcal{W}(\tilde{\bm{r}}_{\bm{p};t}, \textcolor{St}{\mathcal{S}_t})$ (Eq. \ref{eq:bundle_adjust_ray})
    \If {it$|2$} \Comment{\textcolor{StaticNet}{\textbf{Static Net}} and \textcolor{RayRefinement}{\textbf{Ray Warp.}}}
        \State Compute $\hat{\bm{C}}^s(\hat{\bm{r}}_{\bm{p};t})$, $\bm{M}(\hat{\bm{r}}_{\bm{p};t})$
        \State $loss \leftarrow \mathcal{L}_{mphoto} (\hat{\bm{C}}^s(\hat{\bm{r}}_{\bm{p};t}))$ (Eq. \ref{eq:loss_rgb_static})
        \State Update \textcolor{StaticNet}{\textbf{Static Net} $F_{\theta^s}$}, \textcolor{St}{$\mathcal{S}_t$}
    \Else \Comment{\textcolor{DynamicNet}{\textbf{Dynamic Net}} and \textcolor{StaticNet}{\textbf{Static Net}}}
        \State Compute $\hat{\bm{C}}^s(\hat{\bm{r}}_{\bm{p};t}), \hat{\bm{C}}^d(\hat{\bm{r}}_{\bm{p};t})$, $\bm{M}(\hat{\bm{r}}_{\bm{p};t})$
        \State Compute $\hat{\bm{C}}^{full}(\hat{\bm{r}}_{\bm{p};t})$ (Eq. \ref{eq:full_color_rendering})
        \State
        \vspace*{-0.5\baselineskip}
        \begin{equation*}
        \hspace*{2.5\baselineskip}
            \begin{aligned}
                loss \leftarrow &\mathcal{L}_{mphoto} (\hat{\bm{C}}^s(\hat{\bm{r}}_{\bm{p};t})) + \mathcal{L}_{photo}(\hat{\bm{C}}^{d,full}(\hat{\bm{r}}_{\bm{p};t})) \\ + &\mathcal{L}_{sm} + \mathcal{L}_{lg} \quad \text{(Eq. \ref{eq:loss_rgb_dynamic}, Eq. \ref{eq:loss_rgb_static}, Eq. \ref{eq:loss_normal})}
            \end{aligned}
        \end{equation*}
        \State Update \textcolor{StaticNet}{\textbf{Static Net} $F_{\theta^s}$}, \textcolor{DynamicNet}{\textbf{Dynamic Net} $F_{\theta^d}$}
    \EndIf
\EndFor
\EndProcedure
\end{algorithmic}
\label{alg:bri_stage}
\end{algorithm}

\subsection{Motion Decomposition-based Deblurring (MDD) Stage}
\label{sect:deblur_module}
We introduce the MDD stage with a novel Incremental Latent Sharp-rays Prediction (ILSP) approach to predicting latent sharp rays to simulate the motion blur process, as shown in Fig. \ref{fig:MoBluRF}-(b) combined with Algo. \ref{alg:bri_stage} (BRI Stage), which is fully described in Algo. \ref{alg:overall_process} (BRI + MDD Stages). \jm{To accurately predict latent sharp rays under complex camera and object motions during the exposure time, we employ a two-step approach: First, we perform a global estimation of the latent sharp rays, termed Global Motion-aware Ray Prediction (GMRP), by accounting for camera and object motions across both static and dynamic regions; Next, we refine these rays locally and incrementally, focusing specifically on dynamic regions of the scene by incorporating object motions, in a process referred to as Local Object-motion-aware Ray Refinement (LORR).}

\noindent \textbf{\jm{Global Motion-aware Ray Prediction (GMRP).}} 
\jm{Inspired by 2D deblurring methods~\cite{bahat2017non, chakrabarti2016neural, cho2009fast, schuler2015learning, shan2008high, xu2010two, sun2015learning, fang2023self} which approximately predict the projected 2D spatial motion blur kernel from 3D motion for each 2D target pixel, we extend this concept to implicitly estimate a 3D spatial blur kernel that models the blurriness of each target base ray during the exposure time. To model the motion blur process which occurs in both static and dynamic scene components,} we estimate multiple latent sharp rays $\{\dot{\bm{r}}^g_{\bm{p};t;q}\}_{q=1}^{N_b}$ (Fig. \ref{fig:MoBluRF}-(b)) based on the initialized base ray $\hat{\bm{r}}_{\bm{p};t}$ (Eq. \ref{eq:bundle_adjust_ray} and Fig. \ref{fig:MoBluRF}-(a)) as:
\begin{equation}
        \{\dot{\bm{r}}^g_{\bm{p};t;q}\}_{q=1}^{N_b} = \{\mathcal{W}(\hat{\bm{r}}_{\bm{p};t}, \mathcal{S}^g_{t;q})\}_{q=1}^{N_b},
        \label{eq:global_motion_ray}
\end{equation} 
where $\mathcal{W}$ is defined in Eq. \ref{eq:bundle_adjust_ray} and $\mathcal{S}^g_{t;q}$ is a global motion-aware screw axis which is a learnable embedding of $t$ for the $q^{\text{th}}$ latent sharp ray $\dot{\bm{r}}^g_{\bm{p};t;q}$. The GMRP maps the base ray $\hat{\bm{r}}_{\bm{p};t}$ that is the output of the BRI stage to the $N_b$ predicted latent sharp rays $\{\dot{\bm{r}}^g_{\bm{p};t;q}\}_{q=1}^{N_b}$, considering camera and objects motions (one-to-$N_b$ mapping).

\noindent \textbf{\jm{Local Object-motion-aware Ray Refinement (LORR).}} 
\jm{The GMRP can predict latent sharp rays by incorporating both camera and object motions. However, we observed that relying solely on GMRP might lead to instability in deblurring the dynamic radiance field in 3D space, resulting in afterimage artifacts, as illustrated in Fig. \ref{fig:ablation_mdd}-(b). To handle this issue, we introduce LORR as an additional prior to regularize the optimization for more robust estimation of the latent sharp rays under the compound of camera and object motions.} This is accomplished by refining the $q^{\text{th}}$ predicted latent sharp ray $\dot{\bm{r}}^g_{\bm{p};t;q}$ considering pixel-wise local object motion as:
\begin{equation}
        \dot{\bm{r}}^l_{\bm{p};t;q} = \mathcal{W}(\dot{\bm{r}}^g_{\bm{p};t;q}, \mathcal{S}^l_{\bm{p};t;q}),
        \label{eq:local_motion_ray}
\end{equation}
where $\mathcal{S}^l_{\bm{p};t;q} = F_{\theta^l}(\lceil\phi(\dot{\bm{r}}^g_{\bm{p};t;q}), l(t)\rfloor)$ is a local object-motion-aware screw axis learned by a local object-motion MLP $F_{\theta^l}$ which takes a discretized ray embedding $\phi(\dot{\bm{r}}^g_{\bm{p};t;q})$~\cite{PialaC2021termi} and the encoded time $l(t)$ as inputs. $\lceil \cdot \rfloor$ refers to channel-wise concatenation. 
That is, the LORR maps each predicted latent sharp ray $\dot{\bm{r}}^g_{\bm{p};t;q}$ to a single corresponding refined latent sharp ray $\dot{\bm{r}}^l_{\bm{p};t;q}$ considering the local object motion (one-to-one mapping). Specifically, the LORR is only applied to the dynamic scene components which are indicated by the binary motion mask $\bm{M}(\hat{\bm{r}}_{\bm{p};t})=1$.

To obtain the blurry color $\hat{\bm{C}}_\mathcal{B}$, we apply Eq. \ref{eq:avg_color} to predicted latent sharp rays from Eq. \ref{eq:global_motion_ray} for the static scene components or from Eq. \ref{eq:local_motion_ray} for the dynamic scene components. This motion blur process is applied only during the training. In the inference, as shown in Algo. 3, the full color $\hat{\bm{C}}^{full}(\bm{r}^{inf}_{\bm{p};t})$ of target ray $\bm{r}^{inf}_{\bm{p};t}$ is synthesized for sharp target novel view.

\begin{algorithm}
\footnotesize
\caption{Overall Training Process of MoBluRF}
\begin{algorithmic}[1]
\State \textbf{Init} \textcolor{StaticNet}{$F_{\theta^s}$}, \textcolor{DynamicNet}{$F_{\theta^d}$},  \textcolor{Fthetal}{$F_{\theta^l}$}, \textcolor{St}{$\mathcal{S}_t$}, \textcolor{Sgtq}{$\mathcal{S}^g_{t;q}$}
\State \textbf{Do} BRI(\textcolor{StaticNet}{$F_{\theta^s}$}, \textcolor{DynamicNet}{$F_{\theta^d}$}, \textcolor{St}{$\mathcal{S}_t$}) (Fig. \ref{fig:MoBluRF}-(a) and Algo. \ref{alg:bri_stage})
\For{it $=0$ to $10^5$} \Comment{MDD (Sec. \ref{sect:deblur_module} and Fig. \ref{fig:MoBluRF}-(b))}
    \State Sample random rays $\tilde{\bm{r}}_{\bm{p};t}$
    \State Compute base ray $\hat{\bm{r}}_{\bm{p};t} \leftarrow \mathcal{W}(\tilde{\bm{r}}_{\bm{p};t}, \textcolor{St}{\mathcal{S}_t})$ (Eq. \ref{eq:bundle_adjust_ray}) \Comment{Freeze $\textcolor{St}{\mathcal{S}_t}$}
    \State Compute $\hat{\bm{C}}^{s,d,full}(\hat{\bm{r}}_{\bm{p};t}), \bm{M}(\hat{\bm{r}}_{\bm{p};t})$
    \State $        \{\dot{\bm{r}}^g_{\bm{p};t;q}\}_{q=1}^{N_b} \leftarrow \{\mathcal{W}(\hat{\bm{r}}_{\bm{p};t}, \textcolor{Sgtq}{\mathcal{S}^g_{t;q}})\}_{q=1}^{N_b}$ (Eq. \ref{eq:global_motion_ray}) \Comment{GMRP}
    \If {$\bm{M}(\hat{\bm{r}}_{\bm{p};t})$ = 0}
        \For {${q=1}$ to ${N_b}$}
        \State $\dot{\bm{r}}_{\bm{p};t;q} \leftarrow \dot{\bm{r}}^g_{\bm{p};t;q}$
        \State Compute $\hat{\bm{C}}^{s,d,full}(\dot{\bm{r}}_{\bm{p};t;q})$
        \EndFor
    \ElsIf{$\bm{M}(\hat{\bm{r}}_{\bm{p};t})$ = 1}
        \For {${q=1}$ to ${N_b}$}
            \State $\mathcal{S}^l_{\bm{p};t;q} \leftarrow \textcolor{Fthetal}{F_{\theta^l}}(\lceil\phi(\dot{\bm{r}}^g_{\bm{p};t;q}), l(t)\rfloor)$
            \State $\dot{\bm{r}}^l_{\bm{p};t;q} \leftarrow \mathcal{W}(\dot{\bm{r}}^g_{\bm{p};t;q}, \mathcal{S}^l_{\bm{p};t;q})$ (Eq. \ref{eq:local_motion_ray})  \Comment{LORR}
            \State $\dot{\bm{r}}_{\bm{p};t;q} \leftarrow \dot{\bm{r}}^l_{\bm{p};t;q}$
            \State Compute $\hat{\bm{C}}^{s,d,full}(\dot{\bm{r}}_{\bm{p};t;q})$
        \EndFor      
    \EndIf
    \State  
    \vspace*{-0.7\baselineskip}
    \begin{equation*}
    \hspace*{\baselineskip}
    \hat{\bm{C}}_{\mathcal{B}}^{s,d,full}(\hat{\bm{r}}_{\bm{p};t}) \leftarrow \mathcal{A}(\hat{\bm{C}}^{s,d,full}(\hat{\bm{r}}_{\bm{p};t}), \{\hat{\bm{C}}^{s,d,full}(\dot{\bm{r}}_{\bm{p};t;q})\}_{q=1}^{N_b})
    \end{equation*} 
    \State
    \vspace*{-0.7\baselineskip}
    \begin{equation*}
    \hspace*{-2.5\baselineskip}
    \begin{aligned}
        loss \leftarrow & \mathcal{L}_{mphoto} (\hat{\bm{C}}_{\mathcal{B}}^s(\hat{\bm{r}}_{\bm{p};t})) + \mathcal{L}_{photo}(\hat{\bm{C}}_{\mathcal{B}}^{d,full}(\hat{\bm{r}}_{\bm{p};t})) \\ + &\mathcal{L}_{sm} + \mathcal{L}_{lg} \quad \text{(Eq. \ref{eq:loss_rgb_dynamic}, Eq. \ref{eq:loss_rgb_static}, Eq. \ref{eq:loss_normal})}
    \end{aligned}
    \end{equation*}    
    \State Update \textcolor{StaticNet}{\textbf{Static Net} $F_{\theta^s}$}, \textcolor{DynamicNet}
    {\textbf{Dynamic Net} $F_{\theta^d}$}, \textcolor{Fthetal}{$F_{\theta^l}$}, \textcolor{Sgtq}{$\mathcal{S}^g_{t;q}$}
\EndFor
\end{algorithmic}
\label{alg:overall_process}
\end{algorithm}

\begin{algorithm}
\footnotesize
\caption{Inference Process}
\begin{algorithmic}[1]
\Procedure{Inference}{\textcolor{StaticNet}{$F_{\theta^s}$}, \textcolor{DynamicNet}{$F_{\theta^d}$}, $\bm{r}^{inf}_{\bm{p};t}$}
\For{n $=0$ to $128$}
\State Sampling point $\mathbf{x}_n$ on target ray $\bm{r}^{inf}_{\bm{p};t}$ 
\vspace*{0.2\baselineskip}
\State $\bm{c}^s_n, \sigma^s_n, p(\text{st}|\textbf{x}_n) \leftarrow F_{\theta^s}(\gamma(\mathbf{x}_n, \mathbf{d}))$
\vspace*{0.2\baselineskip}
\State $\bm{c}^d_n, \sigma^d_n \leftarrow F_{\theta^d}(\gamma(\mathbf{x}_n, \mathbf{d}), l(t))$
\vspace*{0.2\baselineskip}
\EndFor
\State Compute $\hat{\bm{C}}^{full}(\bm{r}^{inf}_{\bm{p};t})$ (\ref{eq:full_color_rendering})
\EndProcedure
\end{algorithmic}
\label{alg:inference}
\end{algorithm}

\subsection{Loss Functions}
\label{sec:loss}
We use diverse loss functions to train MoBluRF as follows.

\noindent \textbf{Photometric Loss.} We minimize an L2 loss between rendered color $\hat{\bm{C}}(\hat{\bm{r}}_{\bm{p};t})$ and blurry GT color $\bm{\mathcal{B}}_{\bm{p};t}$ as:
\begin{equation}
    \mathcal{L}_{photo}(\hat{\bm{C}}(\hat{\bm{r}}_{\bm{p};t})) = \textstyle\sum_{\hat{\bm{r}}_{\bm{p};t}} \lVert \hat{\bm{C}}(\hat{\bm{r}}_{\bm{p};t}) - \bm{\mathcal{B}}_{\bm{p};t}\rVert^2_2,
\label{eq:loss_rgb_dynamic}
\end{equation}
where $\hat{\bm{C}}(\hat{\bm{r}}_{\bm{p};t})$ can be the rendered color of dynamic or full scene components $\hat{\bm{C}}^{d,full}(\hat{\bm{r}}_{\bm{p};t})$ in the BRI stage and the blurry rendered color $\hat{\bm{C}}_{\mathcal{B}}^{d,full}(\hat{\bm{r}}_{\bm{p};t})$ in the MDD stage.
For both of the rendered color $\hat{\bm{C}}^{s}(\hat{\bm{r}}_{\bm{p};t})$ of static scene components (static regions) in the BRI stage and the blurry rendered color $\hat{\bm{C}}_{\mathcal{B}}^{s}(\hat{\bm{r}}_{\bm{p};t})$ in the MDD stage, we adopt a masked photometric loss to prevent learning the dynamic scene components (dynamic regions) by using $\bm{M}(\hat{\bm{r}}_{\bm{p};t})$ as:
\begin{equation}
\scalebox{0.9}{$
        \mathcal{L}_{mphoto} (\hat{\bm{C}}(\hat{\bm{r}}_{\bm{p};t})) = \textstyle \sum_{\hat{\bm{r}}_{\bm{p};t}} \lVert (\hat{\bm{C}}(\hat{\bm{r}}_{\bm{p};t})) - \bm{\mathcal{B}}_{\bm{p};t}) \cdot (1 - \bm{M}(\hat{\bm{r}}_{\bm{p};t})) \rVert^2_2.
$}
\label{eq:loss_rgb_static}
\end{equation}

\noindent \textbf{Unsupervised Staticness Maximization Loss ($\mathcal{L}_{sm}$).}
We observe that without any regularization for decomposition, $F_{\theta^s}$ and $F_{\theta^d}$ exhibit different convergence speeds, due to multi-view inconsistency for dynamic objects along time evolution in $F_{\theta^s}$ which lacks any time modeling. As a result, $F_{\theta^d}$ quickly surpasses $F_{\theta^s}$ from the early stage, causing $F_{\theta^s}$ to get degenerated, thus causing the staticness probabilities $p(\text{st}|\mathbf{x})$ at 3D positions $\mathbf{x}$ to become all 0. Furthermore, the existing dynamic NeRFs \cite{gao2021dynamic, li2021neural, li2023dynibar, liu2023robust, yan2023nerf, sun2024dyblurf} often rely on off-the-shelf models~\cite{teed2020raft, he2017mask} to obtain the moving object masks for the supervision of decomposition. However, for blurry video frames, it is difficult to obtain accurate motion masks because the off-the-shelf models~\cite{teed2020raft, he2017mask} may fail to consistently capture the accurate segmentation masks of moving objects (Fig. \ref{fig:ablation_sm_loss}). To overcome these drawbacks, our MoBluRF is designed to predict the staticness probability $p(\text{st}|\mathbf{x})$ and the binary motion mask $\bm{M}(\hat{\bm{r}}_{\bm{p};t})$ in an \textit{unsupervised} manner. 
Specifically, we \textit{newly} introduce a simple but effective loss function as an Unsupervised Staticness Maximization loss $\mathcal{L}_{sm}$ that is the sum of absolutes on the logarithms of $p(\text{st}|\mathbf{x})$ weighted by $\lambda_{sm}$, which is denoted as $\mathcal{L}_{sm} = \lambda_{sm}\sum_{\bm{r}_{\bm{p};t}}|\log(p(\text{st}|\mathbf{x}))|$. $\mathcal{L}_{sm}$ encourages the staticness probability $p(\text{st}|\mathbf{x})$ to preserve static characteristics, which can lead to robust decomposition.

\noindent \textbf{Local Geometry Variance Distillation ($\mathcal{L}_{lg}$).} 
Previous prior monocular dynamic NeRFs \cite{li2021neural, li2023dynibar, liu2023robust, sun2024dyblurf} commonly incorporate 2D monocular depth estimators \cite{Ranftl2021dpt, ranftl2020towards} to regularize the geometry of radiance fields. However, as the monocular depths are of scale- and shift-ambiguities, it is harmful to directly supervise them to learn the scene geometries.
NSFF \cite{li2021neural}, RoDynRF \cite{liu2023robust}, and DyBluRF~\cite{sun2024dyblurf} proposed to use the normalized GT depths and the NeRF's rendered depths learned by a scale- and shift-invariant supervision loss. DynIBaR~\cite{li2023dynibar} points out the instability of the above scale- and shift-invariant depth loss and proposes to obtain scale- and shift-consistent depth maps by preprocessing the video sequence using the method in \cite{zhang2021consistent}. However, the consistent depth preprocessing is computationally much more expensive than 2D monocular depth estimation for long video sequences due to the time-consuming optical flow prediction~\cite{teed2020raft} and exhausted keypoints matching. To overcome the above limitations, we propose a novel Local Geometry Variance Distillation to robustly regularize the density field of our $F_{\theta^d}$ with pseudo-GT depths (scale- and shift-ambiguous depths estimated by pretrained models). Specifically, using first-order finite differences, we approximate the gradients of 3D unprojection corresponding to the local kernel of the current pixel $\bm{p}=(p_u,p_v)$. Then, we compute the unit vector from the cross-product of the approximated partial derivatives of geometry to represent the local geometry variance at pixel $\bm{p}$. The predicted unit vector $\hat{\bm{g}}_{\bm{p};t}$ and its pseudo-GT unit vector (local geometry variance) $\bm{g}_{\bm{p};t}$ are computed as:
\begin{equation}
\begin{split}
     &\hat{\bm{g}}_{\bm{p};t} = \overline{\Bigl(\partial \hat{\bm{r}}_{\bm{p};t}(\kappa*)/\partial p_u \times \partial \hat{\bm{r}}_{\bm{p};t}(\kappa*)/\partial p_v}\Bigr), \\
    &{\bm{g}}_{\bm{p};t} = \overline{\Bigl(\partial \hat{\bm{r}}_{\bm{p};t}(\mathcal{\bm{D}}_{\bm{p};t})/\partial p_u \times \partial \hat{\bm{r}}_{\bm{p};t}(\mathcal{\bm{D}}_{\bm{p};t})/\partial p_v}\Bigr),
\end{split}
\label{eq:lg_eq}
\end{equation}
where $\overline{\mathbf{(w)}}$ is the normalization operation of a vector $\mathbf{w}$ and $\kappa* = \sum_{n=1}^{N} \mathcal{T}^d_n \alpha^d_n s^d_n$ is the volume rendered ray distance for the base ray $\hat{\bm{r}}_{\bm{p};t}$ and $\mathcal{\bm{D}}_{\bm{p};t}$ is the corresponding scale- and shift-ambiguous depth value of pixel $\bm{p}$ at time $t$ from the pretrained DPT~\cite{Ranftl2021dpt}. 
It should be noted that the gradient operation and cross-product normalization in Eq. \ref{eq:lg_eq} remove the shift-ambiguities and the scale-ambiguity of $\mathcal{\bm{D}}_{\bm{p};t}$, respectively. Hence, we can regularize the geometry with monocular depths regardless of both the scale- and shift-ambiguities.
Our local geometry variance distillation is the L2 loss between $\hat{\bm{g}}_{\bm{p};t}$ and $\bm{g}_{\bm{p};t}$ weighted by a constant $\lambda_{lg}$ as: 
\begin{equation}
    \mathcal{L}_{lg} = \lambda_{lg}\textstyle\sum_{\bm{p};t} \lVert \hat{\bm{g}}_{\bm{p};t} - \bm{g}_{\bm{p};t}\rVert^2_2.
\label{eq:loss_normal}
\end{equation}

\noindent \jm{Inspired by the scale-invariant depth loss based on gradient matching in MegaDepth \cite{MDLi18}, our proposed $\mathcal{L}_{lg}$ further incorporates the shift-invariance by the supervision of partial gradients through the normalized cross-product. This modification is designed to address both the scale- and shift-ambiguities.}

\section{Experiments}
\label{sec:experiment}
\noindent \textbf{Implementation Details.}
Our MoBluRF is implemented using JAX and built upon Dycheck \cite{gao2022monocular} codebase. Similar to Dycheck \cite{gao2022monocular}, we adopt $N = 128$ samples for each cast ray. Our model undergoes training for $2 \times 10^5$ iterations in the BRI stage and $1 \times 10^5$ in the MDD stage. The hyperparameters for the loss weighting coefficients, denoted as $\lambda_{lg}$ and $\lambda_{sm}$, are empirically set to 0.075 and 0.002, respectively. For all scenes, our best model is configured with a fixed number of latent sharp rays $N_b=6$. We utilize the Adam optimizer with an exponential scheduling for training our networks. Notably, the learning rates for the screw axis embeddings $\mathcal{S}_t$ and $\mathcal{S}^g_{t;q}$ are in the range $[10^{-4}, 10^{-6}]$, while the learning rates for the remaining MLPs are in the range $[10^{-3}, 10^{-4}]$.
For the implementation of the Static Net $F_{\theta^s}$ and Dynamic Net $F_{\theta^d}$, we adopt the NeRF architecture from the Dycheck \cite{gao2022monocular} codebase to ensure a fair comparison. This NeRF architecture comprises a trunk MLP, a sigma MLP and an rgb MLP. The sigma and rgb MLPs take the output feature of the trunk MLP as their respective inputs. We set the numbers of layers for the trunk, sigma, and rgb MLPs to 9, 1, and 2, respectively. The dimensions of the hidden layers for the trunk and rgb MLPs are set to 256 and 128, respectively. For $F_{\theta^s}$, we introduce a blending MLP that consists of a single layer. This MLP takes the output of the trunk MLP as input to produce the staticness probability $p(\text{st}|\textbf{x})$. For the local object-motion MLP $F_{\theta^l}$, we set the number of layers and the dimension of hidden layers to 8 and 128, respectively. We set the number of frequencies for each positional encoding $\gamma$ of 3D sample positions $\textbf{x}$ and their corresponding viewing directions $\textbf{d}$ to 8 and 4, respectively. We also set the output dimension of the time encoding $l(t)$ to 8 and the number of samples for the discretized ray embedding $\phi$ to 32.

\noindent \textbf{Datasets.}
\jm{To evaluate the performance of MoBluRF and other methods in deblurring dynamic novel view synthesis, we use the following three datasets.}

\begin{figure}[h]
    \centering
    \includegraphics[scale=0.15]{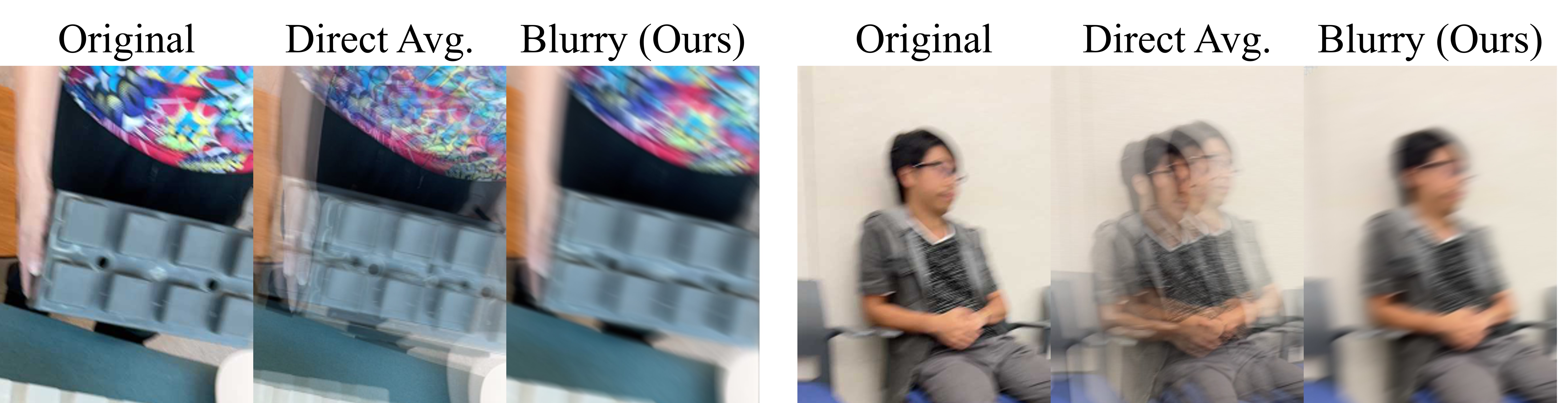}
    \vspace{-6mm}
    \caption{Visualizations of blurry frame synthesis. Our synthesized blurry frames using the VFI method~\cite{sim2021xvfi} exhibit more realistic and natural blurriness compared to the average of three consecutive original frames (`Direct Avg.').}
    \label{fig:synthesis_blurry}
\end{figure}

\textit{1) Blurry iPhone Dataset:} We propose a new blurry version of the iPhone dataset~\cite{gao2022monocular}, referred to as the Blurry iPhone dataset. This dataset comprises 7 video sequences, each containing an average length of 365 frames. The videos feature a wide range of complex object motions, including rotations, hand movements, and rapidly changing actions, complemented by swift camera movements~\cite{gao2022monocular}. It includes synthetic blurry video frames $\{\bm{\mathcal{B}}_{t}\}_{t=1}^{N_f}$ along with their corresponding camera poses $\{\tilde{\bm{\mathcal{P}}}_{t}\}_{t=1}^{N_f}$ for training, and includes the original video frames and camera poses of the iPhone dataset~\cite{gao2022monocular} for evaluation. To synthesize a blurry video frame $\bm{\mathcal{B}}_t$, we follow the standard video blur synthesis process~\cite{sun2024dyblurf, nah2017deep, Nah_2019_CVPR_Workshops_REDS}. We apply the VFI method~\cite{sim2021xvfi} to the original iPhone dataset~\cite{gao2022monocular} to increase the frame rate by 8$\times$. The interpolated frames are then averaged with the original center frame to generate each final blurry frame $\bm{\mathcal{B}}_t$. Fig. \ref{fig:synthesis_blurry} visualizes some synthesized blurry frames and compares them with the direct averaging of three original consecutive frames, without applying the VFI method~\cite{sim2021xvfi}. As shown, the synthesis process of our Blurry iPhone dataset results in more natural and realistic blurriness compared to the direct averaging approach.

\begin{table*}
\caption{motion deblurring novel view synthesis evaluation on the Blurry iPhone dataset. \textcolor{red}{\textbf{Red}} and \textcolor{blue}{\underline{blue}} denote the best and second best performances, respectively. Each block element of 4-performance denotes (\textbf{mPSNR$\uparrow$ / mSSIM$\uparrow$ / mLPIPS$\downarrow$ / tOF$\downarrow$}).}
\vspace{-3mm}
\begin{center}
\setlength\tabcolsep{6pt} 
\renewcommand{\arraystretch}{1.1}
\scalebox{0.8}{
\begin{tabular}{ c | c c c c }
\bottomrule
\hline\noalign{\smallskip}
Methods & Apple & Block & Paper-windmill & Space-out \\  
\bottomrule
\hline\noalign{\smallskip}
TiNeuVox~\cite{fang2022fast}       & 13.53 / 0.680 / 0.723 / 1.704 & 10.79 / 0.558 / 0.676 / 1.705 & 14.15 / 0.273 / 0.781 / 4.108 & 14.18 / 0.557 / 0.587 / 1.385  \\
HexPlane~\cite{cao2023hexplane}    & 16.80 / 0.715 / \textcolor{blue}{\underline{0.523}} / 1.239 & 15.58 / 0.604 / 0.459 / 0.820 & \textcolor{blue}{\underline{17.11}} / \textcolor{blue}{\underline{0.352}} / 0.422 / \textcolor{blue}{\underline{0.318}} & 14.73 / 0.558 / 0.511 / 1.270 \\
T-NeRF~\cite{gao2022monocular}   & \textcolor{blue}{\underline{17.34}} / \textcolor{blue}{\underline{0.720}} / 0.547 / \textcolor{blue}{\underline{0.620}} & \textcolor{blue}{\underline{16.48}} / \textcolor{blue}{\underline{0.644}} / 0.411 / \textcolor{blue}{\underline{0.795}}     & 16.83 / 0.338 / 0.424 / 0.569     & 16.23 / 0.561 / 0.436 / 1.329     \\
HyperNeRF~\cite{park2021hypernerf} & 14.31 / 0.681 / 0.663 / 1.411     & 16.12 / 0.642 / 0.416 / 0.958     & 16.59 / 0.335 / 0.365 / 0.666     & \textcolor{blue}{\underline{17.79}} / \textcolor{blue}{\underline{0.631}} / 0.332 / 0.402     \\
4D-GS~\cite{wu20234d}       & 14.71 / 0.680 / 0.626 / 1.105 & 12.19 / 0.523 / 0.619 / 1.243 & 14.45 / 0.226 / 0.484 / 0.546 & 14.91 / 0.533 / 0.427 / \textcolor{blue}{\underline{0.397}}  \\
\midrule
DP-NeRF$_t$~\cite{lee2023dp}           & 11.97 / 0.665 / 0.717 / 2.072 &  9.96 / 0.514 / 0.729 / 1.602 & 12.66 / 0.241 / 0.713 / 1.482 & 13.15 / 0.532 / 0.628 / 0.639 \\
BAD-NeRF$_t$~\cite{wang2023bad}        & 12.29 / 0.668 / 0.744 / 1.743 &  9.61 / 0.517 / 0.736 / 1.290 & 12.44 / 0.266 / 0.564 / 0.973 & 12.57 / 0.508 / 0.643 / 0.437 \\
DyBluRF~\cite{sun2024dyblurf}        & 13.13 / 0.670 / 0.702 / 0.903 &  10.74 / 0.506 / 0.722 / 1.602 & 12.77 / 0.241 / 0.752 / 2.217 & 13.91 / 0.520 / 0.670 / 2.202 \\
\midrule
GShiftNet~\cite{li2023simple} + \cite{gao2022monocular}       & 16.83 / 0.719 / 0.532 / 0.666 & 16.32 / 0.641 / \textcolor{blue}{\underline{0.403}} / 0.839 & 16.65 / 0.323 / 0.369 / 0.476 & 17.75 / 0.617 / 0.347 / 0.455  \\
GShiftNet~\cite{li2023simple} + \cite{park2021hypernerf}       & 14.31 / 0.676 / 0.622 / 1.332 & 15.71 / 0.626 / 0.417 / 0.942 & 16.31 / 0.328 / \textcolor{blue}{\underline{0.310}} / 0.566 & 17.52 / 0.618 / \textcolor{red}{\textbf{0.296}} / 0.555  \\
GShiftNet~\cite{li2023simple} + \cite{wu20234d}       & 14.62 / 0.674 / 0.619 / 1.111 & 12.46 / 0.525 / 0.623 / 0.950 & 14.35 / 0.214 / 0.468 / 0.643 & 14.71 / 0.524 / 0.401 / 0.412  \\
\midrule
\textbf{MoBluRF (Ours)}            & \textcolor{red}{\textbf{18.03}} / \textcolor{red}{\textbf{0.737}} / \textcolor{red}{\textbf{0.505}} / \textcolor{red}{\textbf{0.479}} & \textcolor{red}{\textbf{17.35}} / \textcolor{red}{\textbf{0.660}} / \textcolor{red}{\textbf{0.361}} / \textcolor{red}{\textbf{0.735}} & \textcolor{red}{\textbf{18.08}} / \textcolor{red}{\textbf{0.413}} / \textcolor{red}{\textbf{0.282}} / \textcolor{red}{\textbf{0.238}} & \textcolor{red}{\textbf{18.83}} / \textcolor{red}{\textbf{0.644}} / \textcolor{blue}{\underline{0.314}} / \textcolor{red}{\textbf{0.264}}
\end{tabular}
}
\scalebox{0.8}{
\begin{tabular}{ c | c c c | c }
\bottomrule
\hline\noalign{\smallskip}
Methods & Spin & Teddy & Wheel & \textbf{Average} \\  
\bottomrule
\hline\noalign{\smallskip}
TiNeuVox~\cite{fang2022fast}       & 11.13 / 0.411 / 0.726 / 2.239 & 10.28 / 0.496 / 0.834 / 1.304 & 9.48  / 0.312 / 0.717 / 3.556 & 11.93 / 0.470 / 0.721 / 2.286  \\
HexPlane~\cite{cao2023hexplane}    & 16.02 / 0.482 / 0.563 / 1.253 & 12.84 / 0.497 / 0.587 / 1.220 & 12.87 / 0.409 / 0.521 / \textcolor{blue}{\underline{1.336}} & 15.14 / 0.517 / 0.512 / 1.065  \\
T-NeRF~\cite{gao2022monocular}   & \textcolor{blue}{\underline{17.16}} / \textcolor{blue}{\underline{0.503}} / 0.534 / \textcolor{blue}{\underline{1.162}} & \textcolor{blue}{\underline{14.07}} / 0.562 / 0.464 / \textcolor{blue}{\underline{1.094}} & 14.93 / 0.499 / 0.379 / 1.360 & 16.15 / 0.547 / 0.456 / 0.990 \\
HyperNeRF~\cite{park2021hypernerf} & 16.39 / 0.498 / 0.499 / 1.277 & 13.77 / \textcolor{blue}{\underline{0.567}} / \textcolor{red}{\textbf{0.420}} / 1.143 & 12.11 / 0.393 / 0.435 / 1.739 & 15.30 / 0.535 / 0.447 / 1.085     \\
4D-GS~\cite{wu20234d}       & 14.31 / 0.418 / 0.513 / 1.522 & 12.18 / 0.492 / 0.564 / 1.402 & 10.44 / 0.309 / 0.583 / 2.004 & 13.31 / 0.455 / 0.545 / 1.174  \\
\midrule
DP-NeRF$_t$~\cite{lee2023dp}           & 10.65 / 0.404 / 0.730 / 1.956 & 10.40 / 0.480 / 0.760 / 1.482 &  9.26 / 0.299 / 0.736 / 2.561 & 11.15 / 0.448 / 0.716 / 1.685  \\
BAD-NeRF$_t$~\cite{wang2023bad}        & 10.59 / 0.404 / 0.741 / 1.722 &  9.77 / 0.457 / 0.758 / 1.537 &  9.23 / 0.303 / 0.748 / 2.544 & 10.93 / 0.446 / 0.705 / 1.464  \\
DyBluRF~\cite{sun2024dyblurf}        & 11.06 / 0.395 / 0.702 / 3.440 &  9.09 / 0.464 / 0.749 / 2.796 & 10.82 / 0.332 / 0.777 / 4.018 & 11.65 / 0.447 / 0.725 / 2.454 \\
\midrule
GShiftNet~\cite{li2023simple} + \cite{gao2022monocular}       & 16.78 / 0.494 / 0.529 / 1.276 & 13.74 / 0.552 / 0.461 / 1.271 & \textcolor{blue}{\underline{15.48}} / \textcolor{blue}{\underline{0.537}} / \textcolor{blue}{\underline{0.310}} / 1.442 & \textcolor{blue}{\underline{16.22}} / \textcolor{blue}{\underline{0.555}} / 0.422 / \textcolor{blue}{\underline{0.918}}  \\
GShiftNet~\cite{li2023simple} + \cite{park2021hypernerf}       & 15.74 / 0.488 / \textcolor{blue}{\underline{0.476}} / 1.488 & 13.66 / 0.566 / \textcolor{blue}{\underline{0.423}} / 1.268 & 10.96 / 0.347 / 0.456 / 2.114 & 14.89 / 0.521 / \textcolor{blue}{\underline{0.382}} / 1.181  \\
GShiftNet~\cite{li2023simple} + \cite{wu20234d}       & 14.46 / 0.411 / 0.502 / 1.572 & 12.13 / 0.500 / 0.547 / 1.343 & 10.29 / 0.306 / 0.577 / 1.954 & 13.29 / 0.450 / 0.534 / 1.141  \\
\midrule
\textbf{MoBluRF (Ours)}
& \textcolor{red}{\textbf{18.20}} / \textcolor{red}{\textbf{0.534}} / \textcolor{red}{\textbf{0.415}} / \textcolor{red}{\textbf{1.011}}
& \textcolor{red}{\textbf{14.61}} / \textcolor{red}{\textbf{0.569}} / 0.451 / \textcolor{red}{\textbf{0.855}}
& \textcolor{red}{\textbf{16.17}} / \textcolor{red}{\textbf{0.579}} / \textcolor{red}{\textbf{0.304}} / \textcolor{red}{\textbf{1.142}}
& \textcolor{red}{\textbf{17.33}} / \textcolor{red}{\textbf{0.591}} / \textcolor{red}{\textbf{0.376}} / \textcolor{red}{\textbf{0.675}} \\
\bottomrule
\hline\noalign{\smallskip}
\end{tabular}
}
\vspace{-0.3cm}
\label{table:quantitative_comparison}
\end{center}
\end{table*}

\begin{figure*}[h]
\centering
\includegraphics[scale=0.3]{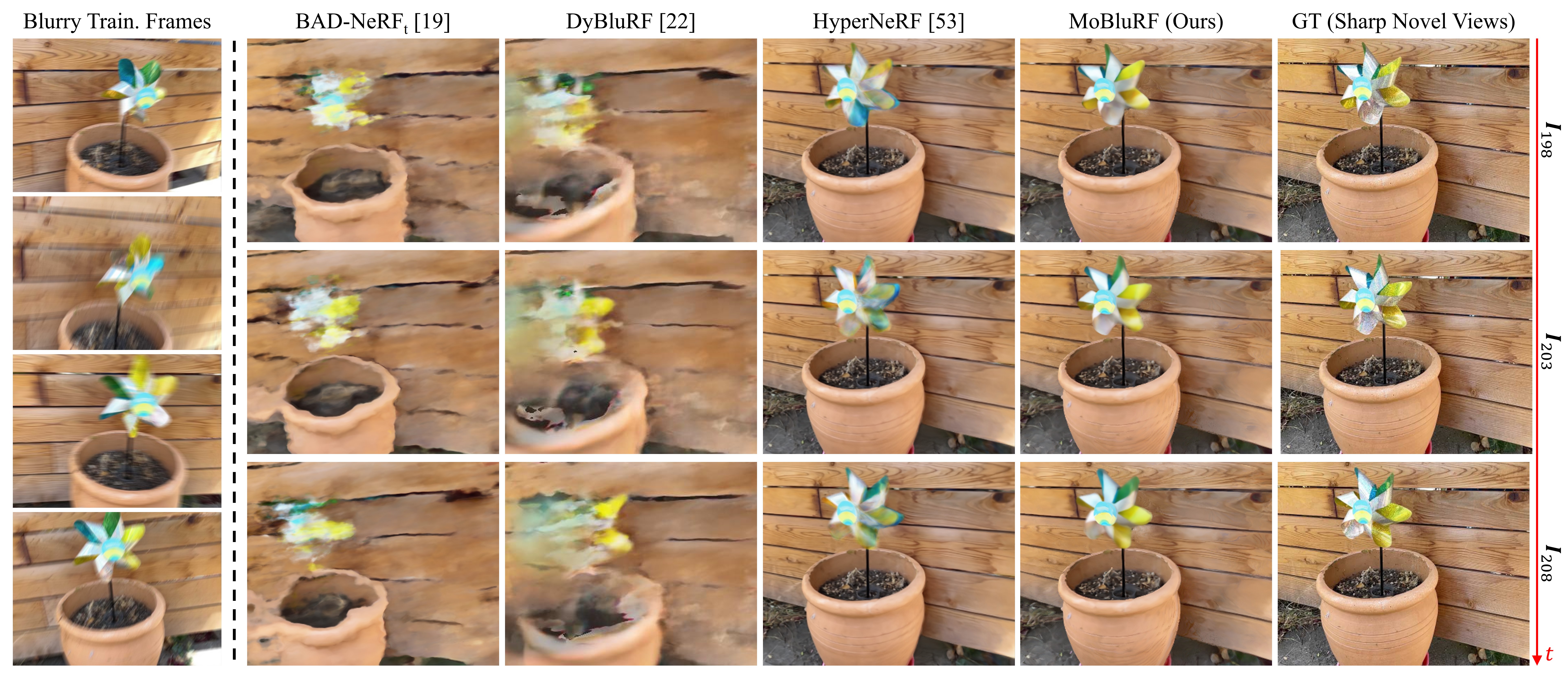}
\vspace{-0.3cm}
\caption{Motion deblurring novel view synthesis results on the Blurry iPhone dataset. We compare the novel view synthesis quality of our MoBluRF for the \textit{Paper-windmill} scene with the existing methods~\cite{wang2023bad, park2021hypernerf, sun2024dyblurf}. Each row corresponds to the 198$^\text{th}$, 203$^\text{th}$ and 208$^\text{th}$ frame, respectively.}
\label{fig:figure_qualitative}
\end{figure*}

Accompanying each blurry frame in the training dataset, we also synthesize the corresponding inaccurate camera pose $\tilde{\bm{\mathcal{P}}}_t=[\tilde{\bm{\mathcal{R}}}_t|\tilde{\bm{\mathcal{T}}}_t]$ by simulating a smooth camera trajectory in 3D space~\cite{argaw2021restoration, kwak2023vivid}. More specifically, the method in~\cite{argaw2021restoration} utilizes a quaternion spherical linear interpolation (Slerp) approach to synthesize intermediate camera poses between two specific camera poses. This process generates the corresponding intermediate images required for the blur modeling process. Following this methodology~\cite{argaw2021restoration} for synthesizing the rotation matrix $\tilde{\bm{\mathcal{R}}}_t$, we generate 8$\times$ interpolated rotation matrices between each pair of original rotation matrices using Slerp. The interpolated matrices are then averaged with the original center matrix to obtain the rotation matrix used for the corresponding blurry frame $\bm{\mathcal{B}}_t$. Since directly calculating the average of multiple rotation matrices is challenging, we convert these matrices into quaternions before averaging them. The resultant $\tilde{\bm{\mathcal{R}}}_t$ appropriately simulates the 3D camera motion during the blur modeling process~\cite{argaw2021restoration, kwak2023vivid}. Similar to the process with the rotation matrix $\tilde{\bm{\mathcal{R}}}_t$ for the synthesis of the translation matrix $\tilde{\bm{\mathcal{T}}}_t$, we use linear interpolation to generate the interpolated matrices, which are then averaged with the original center matrix. The resulting interpolated camera pose $\tilde{\bm{\mathcal{P}}}_t$ has different rotation and translation matrices against the original camera pose $\bm{\mathcal{P}}_t$, making it more challenging for the existing methods~\cite{park2021hypernerf, wang2023bad, lee2023dp, cao2023hexplane, fang2022fast, mildenhall2020nerf, sun2024dyblurf} to reconstruct the correct scene geometry, except our MoBluRF.

\textit{2) Stereo Blur Dataset:} \jm{Stereo Blur dataset~\cite{zhou2019davanet} comprises six scenes with significant motion blur, each containing left-view blurry stereo videos and their corresponding sharp right-view videos. To synthesize motion blur, the dataset is captured using a ZED stereo camera capable of recording dynamic scenes at 60fps. The frame rate is then increased to 480fps using a VFI method~\cite{niklaus2017video}, followed by frame averaging. For each scene, 24 frames are extracted from the original video, and COLMAP~\cite{schonberger2016structure} is used to obtain the camera parameters.}

\textit{3) DAVIS Dataset:} \jm{DAVIS dataset~\cite{pont20172017}, widely used for qualitative comparisons in monocular dynamic novel view synthesis under real-world scenarios~\cite{liu2023robust, stearns2024dynamic, chu2024dreamscene4d, wang2024shape, lei2024mosca}, consists of various in-the-wild monocular videos, with an average of 69.7 frames per scene. This dataset includes several scenes with rapid camera and object movements, exhibiting significant real-world blurriness across video frames. As noted in \cite{liu2023robust}, COLMAP~\cite{schonberger2016structure} fails to produce reasonable camera poses for this real-world dataset~\cite{pont20172017}. Therefore, we adopt the strategy proposed in \cite{wang2024shape} to obtain the camera parameters.}

\noindent \textbf{Metrics.} For comparison of our MoBluRF with other SOTA methods on our proposed Blurry iPhone dataset in the monocular video settings, we utilize the co-visibility masked image metrics, including mPSNR, mSSIM, and mLPIPs, following the approach introduced by Dycheck~\cite{gao2022monocular}. These metrics mask out the regions of the test video frames which are not observed in the training frames. We further utilize tOF~\cite{chu2020learning} to measure the temporal consistency of hundreds of reconstructed video frames. \jm{For comparisons on the Stereo Blur dataset~\cite{zhou2019davanet}, we follow DyBluRF~\cite{sun2024dyblurf} and evaluate each method in terms of PSNR, SSIM, and LPIPS.}

\begin{table*}
\caption{motion deblurring novel view synthesis evaluation on the stereo blur dataset~\cite{zhou2019davanet}. \textcolor{red}{\textbf{Red}} and \textcolor{blue}{\underline{blue}} denote the best and second best performances, respectively. Each block element of 3-performance denotes (\textbf{PSNR$\uparrow$ / SSIM$\uparrow$ / LPIPS$\downarrow$}).}
\vspace{-3mm}
\begin{center}
\setlength\tabcolsep{6pt} 
\renewcommand{\arraystretch}{1.1}
\scalebox{0.77}{
\begin{tabular}{ c | c c c c c c | c }
\bottomrule
\hline\noalign{\smallskip}
Methods & Sailor & Seesaw & Street & Children & Skating & Basketball & Average \\  
\bottomrule
\hline\noalign{\smallskip}
BAD-NeRF~\cite{wang2023bad}       & 16.96 / 0.631 / 0.333 & 20.58 / 0.795 / 0.220 & 20.27 / 0.670 / 0.190 & 18.10 / 0.650 / 0.386 & 19.08 / 0.691 / 0.345 & 17.99 / 0.731 / 0.271 & 18.83 / 0.695 / 0.291  \\
HyperNeRF~\cite{park2021hypernerf} & 18.56 / 0.743 / 0.275 & 20.25 / 0.779 / 0.182 & 19.99 / 0.662 / 0.137 & 21.36 / 0.762 / 0.279 & 19.52 / 0.702 / 0.319 & 21.21 / 0.818 / 0.136 & 20.15 / 0.744 / 0.221 \\
DynamicNeRF~\cite{gao2021dynamic}       & 22.32 / \textcolor{blue}{\underline{0.810}} / 0.247 & 18.14 / 0.775 / 0.158 & 19.06 / 0.669 / 0.176 & 24.00 / 0.823 / 0.307 & 26.18 / 0.907 / 0.124 & 24.10 / 0.880 / 0.143 & 22.30 / 0.811 / 0.193 \\
NSFF~\cite{li2021neural} & 19.06 / 0.726 / 0.290 & 19.92 / 0.807 / 0.178 & 23.42 / 0.800 / 0.121 & \textcolor{blue}{\underline{24.55}} / 0.846 / 0.259 & \textcolor{red}{\textbf{27.96}} / \textcolor{red}{\textbf{0.923}} / 0.116 & \textcolor{blue}{\underline{25.06}} / \textcolor{blue}{\underline{0.903}} / 0.122 & 23.33 / 0.834 / 0.181 \\
RoDynRF~\cite{liu2023robust} & 12.69 / 0.584 / 0.317 & 23.71 / \textcolor{blue}{\underline{0.894}} / 0.116 & 19.80 / 0.719 / 0.161 & 15.02 / 0.609 / 0.382 & 21.66 / 0.831 / 0.160 & 22.82 / 0.850 / 0.166 & 19.28 / 0.748 / 0.217 \\
Restormer~\cite{zamir2022restormer} + \cite{li2021neural}       & 18.96 / 0.728 / 0.252 & 20.19 / 0.827 / 0.149 & 23.39 / 0.790 / 0.125 & \textcolor{red}{\textbf{24.88}} / 0.864 / 0.182 & 26.91 / \textcolor{blue}{\underline{0.922}} / 0.067 & 24.70 / \textcolor{blue}{\underline{0.903}} / 0.084 & 23.17 / 0.839 / 0.143 \\
Restormer~\cite{zamir2022restormer} + \cite{liu2023robust}       & 17.16 / 0.703 / 0.227 & 23.63 / 0.892 / 0.090 & 19.75 / 0.727 / 0.149 & 17.41 / 0.665 / 0.282 & 23.22 / 0.869 / 0.110 & 23.38 / 0.868 / 0.114 & 20.76 / 0.787 / 0.162 \\
JCD~\cite{zhong2021towards} + \cite{li2021neural}       & 22.60 / 0.805 / 0.150 & 21.31 / 0.856 / 0.135 & 23.06 / 0.791 / 0.114 & 24.27 / 0.853 / 0.195 & 26.88 / 0.917 / 0.078 & 24.64 / \textcolor{blue}{\underline{0.903}} / 0.076 & 23.79 / 0.854 / 0.125 \\
JCD~\cite{zhong2021towards} + \cite{liu2023robust}       & 17.28 / 0.772 / 0.225 & 23.85 / 0.893 / 0.080 & 19.69 / 0.700 / 0.153 & 18.39 / 0.683 / 0.226 & 22.54 / 0.849 / 0.126 & 23.51 / 0.867 / 0.098 & 20.88 / 0.794 / 0.151\\
DyBluRF~\cite{sun2024dyblurf}        & \textcolor{blue}{\underline{23.57}} / \textcolor{red}{\textbf{0.863}} / \textcolor{blue}{\underline{0.116}} & \textcolor{blue}{\underline{24.61}} / 0.881 / \textcolor{blue}{\underline{0.072}} & \textcolor{blue}{\underline{26.09}} / \textcolor{red}{\textbf{0.893}} / \textcolor{red}{\textbf{0.069}} & 24.27 / \textcolor{red}{\textbf{0.888}} / \textcolor{blue}{\underline{0.102}} & 27.03 / 0.902 / \textcolor{red}{\textbf{0.077}} & \textcolor{red}{\textbf{25.27}} / \textcolor{red}{\textbf{0.916}} / \textcolor{red}{\textbf{0.053}} & \textcolor{blue}{\underline{25.14}} / \textcolor{blue}{\underline{0.891}} / \textcolor{blue}{\underline{0.082}} \\
\midrule
\textbf{MoBluRF (Ours)}            & \textcolor{red}{\textbf{24.08}} / \textcolor{red}{\textbf{0.863}} / \textcolor{red}{\textbf{0.115}} & \textcolor{red}{\textbf{28.14}} / \textcolor{red}{\textbf{0.944}} / \textcolor{red}{\textbf{0.042}} & \textcolor{red}{\textbf{26.21}} / \textcolor{blue}{\underline{0.870}} / \textcolor{blue}{\underline{0.071}} & 23.78 / \textcolor{blue}{\underline{0.865}} / \textcolor{red}{\textbf{0.095}} & \textcolor{blue}{\underline{27.64}} / \textcolor{blue}{\underline{0.922}} / \textcolor{blue}{\underline{0.080}} & 24.28 / 0.891 / \textcolor{blue}{\underline{0.066}} & \textcolor{red}{\textbf{25.69}} / \textcolor{red}{\textbf{0.893}} / \textcolor{red}{\textbf{0.078}} \\
\bottomrule
\hline\noalign{\smallskip}
\end{tabular}
}
\label{table:quantitative_comparison_stereo}
\end{center}
\end{table*}

\begin{figure*}
\centering
\includegraphics[width=\textwidth]{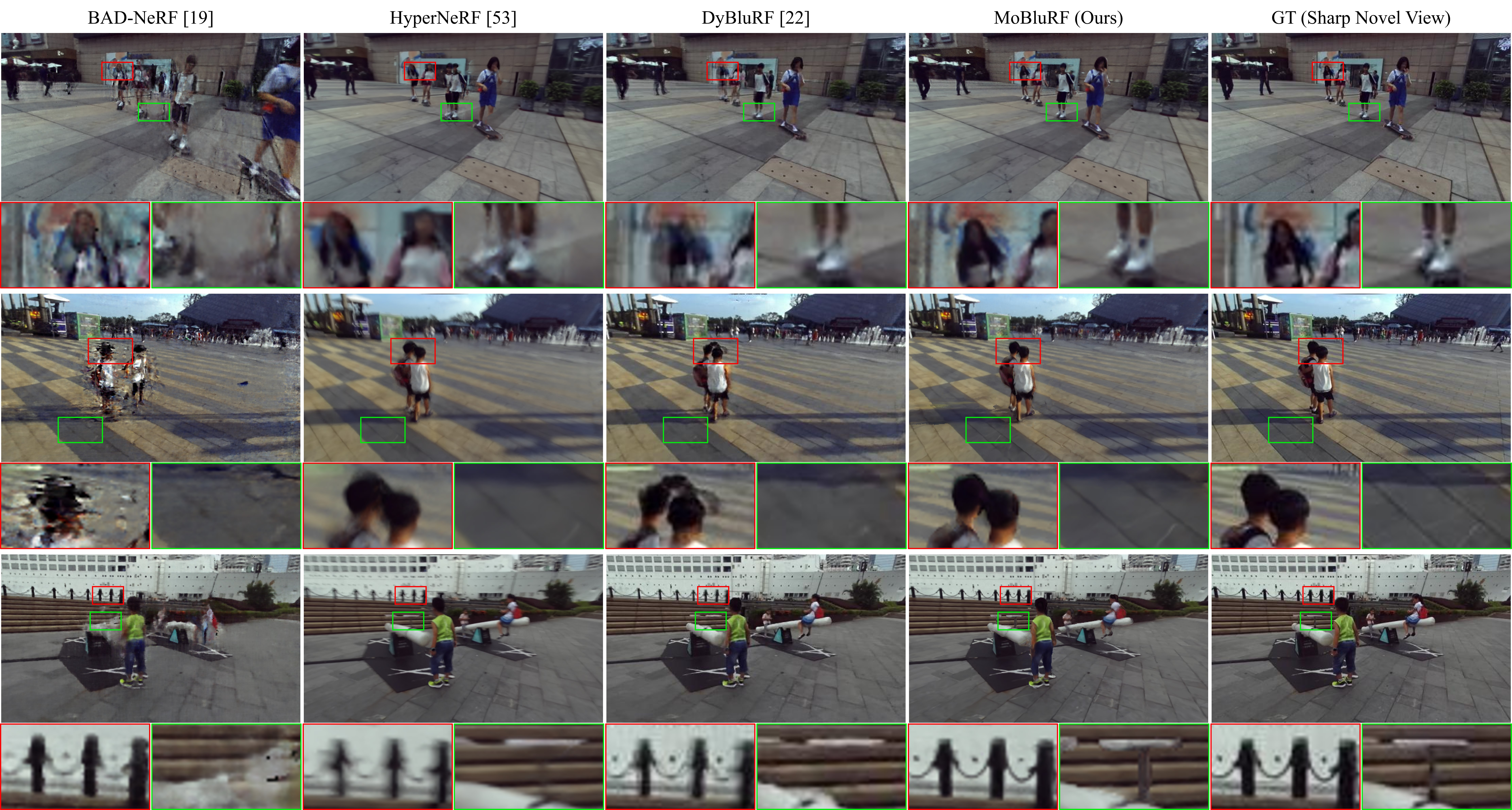}
\caption{\jm{Motion deblurring novel view synthesis results on the Stereo Blur dataset~\cite{zhou2019davanet}. We compare the novel view synthesis quality of our MoBluRF for the \textit{skating}, \textit{children} and \textit{seesaw} scenes with the existing methods~\cite{wang2023bad, park2021hypernerf, sun2024dyblurf}. Best viewed when zoomed in for details.}}
\label{fig:figure_qualitative_dyblurf}
\end{figure*}

\subsection{Experimental Results}
\vspace{-1mm}
We evaluate MoBluRF model from two perspectives. First, we validate the quality of monocular video view synthesis from blurry videos using MoBluRF by comparing it with the SOTA novel view synthesis methods across three datasets. Second, we demonstrate MoBluRF's robustness in dynamic novel view synthesis against the degrees of blurriness by comparing its performance with the existing dynamic novel view synthesis methods~\cite{gao2022monocular, park2021hypernerf} that are trained with either the original iPhone dataset~\cite{gao2022monocular} or our new Blurry iPhone dataset.

\subsubsection{Motion Deblurring Novel View Synthesis} 
\begin{figure*}
\centering
\includegraphics[width=\textwidth]{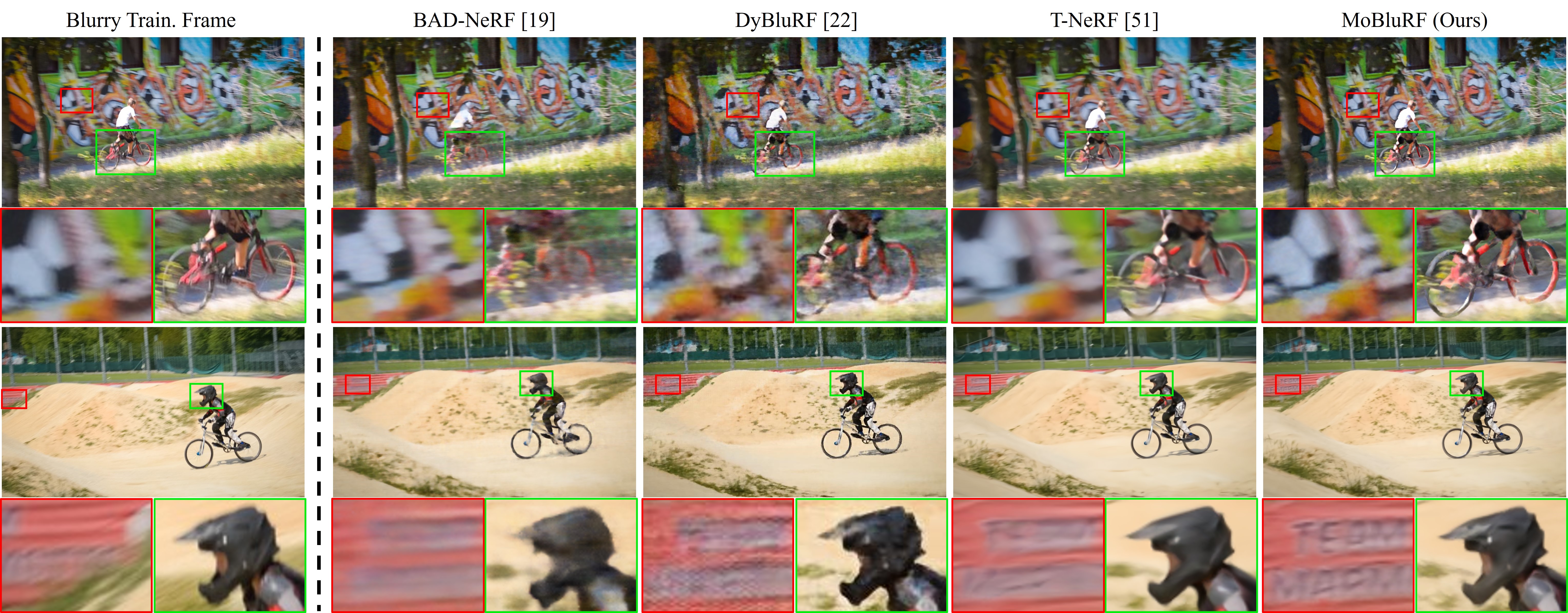}
\vspace{-0.5cm}
\caption{\jm{Motion deblurring novel view synthesis results on the DAVIS dataset~\cite{pont20172017}. We compare the novel view synthesis quality of our MoBluRF for the \textit{bmx-trees} and \textit{bmx-bumps} scenes with the existing methods~\cite{wang2023bad, gao2022monocular, sun2024dyblurf}. Best viewed when zoomed in for details.}}
\label{fig:figure_qualitative_davis}
\end{figure*}

Table \ref{table:quantitative_comparison} shows the effectiveness of MoBluRF in motion deblurring novel view synthesis in comparison with TiNeuVox~\cite{fang2022fast}, HexPlane~\cite{cao2023hexplane}, T-NeRF~\cite{gao2022monocular}, HyperNeRF~\cite{park2021hypernerf}, 4D-GS~\cite{wu20234d}, DP-NeRF~\cite{lee2023dp}, BAD-NeRF~\cite{wang2023bad}, and DyBluRF \cite{sun2024dyblurf} that are all optimized on our Blurry iPhone dataset. For the deblurring methods~\cite{lee2023dp, wang2023bad} initially designed for static scenes, we add time embedding to their network inputs as the same as T-NeRF~\cite{gao2022monocular} in Table \ref{table:quantitative_comparison}, resulting in BAD-NeRF$_t$ and DP-NeRF$_t$, enabling dynamic component synthesis for a fair comparison.
Additionally, we assess our MoBluRF against a cascade approach that combines the SOTA 2D video deblurring method (GShiftNet~\cite{li2023simple}) with each of dynamic NVS methods (T-NeRF~\cite{gao2022monocular}, HyperNeRF~\cite{park2021hypernerf} and 4D-GS~\cite{wu20234d}). Note that for each of these cascades, we train it with the original camera poses from~\cite{gao2022monocular}, bypassing our synthetic camera poses for a fair comparison.
As detailed in Table \ref{table:quantitative_comparison}, our MoBluRF significantly outperforms the existing SOTA methods across all metrics, especially demonstrating superior perceptual quality (mLPIPS) and temporal consistency (tOF). 
Using GShiftNet~\cite{li2023simple} as preprocessing for subsequent NeRF methods~\cite{gao2022monocular,park2021hypernerf,wu20234d} improves perceptual quality (mLPIPS), but leads to more unstable structural qualities (mPSNR, mSSIM) and temporal consistency (tOF). This implies that using 2D pixel-domain deblurring as preprocessing for subsequent dynamic NeRFs may be suboptimal. Therefore, these cascades exhibit significantly poorer performance compared to our MoBluRF.
\jm{Table \ref{table:quantitative_comparison_stereo} demonstrates the effectiveness of MoBluRF in motion deblurring novel view synthesis, compared to BAD-NeRF~\cite{wang2023bad}, HyperNeRF~\cite{park2021hypernerf}, DynamicNeRF~\cite{gao2021dynamic}, NSFF~\cite{li2021neural}, RoDynRF~\cite{liu2023robust}, cascades of 2D deblurring methods (either Restormer~\cite{zamir2022restormer} or JCD~\cite{zhong2021towards}), dynamic NeRF methods (either NSFF~\cite{li2021neural} or RoDynRF~\cite{liu2023robust}), and DyBluRF~\cite{sun2024dyblurf} on the Stereo Blur dataset~\cite{zhou2019davanet}. As shown in Table \ref{table:quantitative_comparison_stereo}, our MoBluRF outperforms SOTA methods in dynamic deblurring novel view synthesis in terms of averaged PSNR, SSIM, and LPIPS on the Stereo Blur dataset~\cite{zhou2019davanet}.}
Fig. \ref{fig:figure_page1} and Fig. \ref{fig:figure_qualitative} demonstrate the visual superiority of our MoBluRF compared to BAD-NeRF~\cite{wang2023bad}, HyperNeRF~\cite{park2021hypernerf} and DyBluRF~\cite{sun2024dyblurf} on our Blurry iPhone dataset. As shown in these figures, our MoBluRF effectively renders sharp novel views even when optimized with blurry training frames, accurately capturing the scene geometry. In contrast, other deblurring NeRF methods~\cite{wang2023bad, sun2024dyblurf} fail to reconstruct the scene geometry correctly. \jm{Fig.~\ref{fig:figure_qualitative_dyblurf} provides qualitative comparisons of our MoBluRF with BAD-NeRF~\cite{wang2023bad}, HyperNeRF~\cite{park2021hypernerf}, and DyBluRF~\cite{sun2024dyblurf} on the Stereo Blur dataset~\cite{zhou2019davanet}.  Fig.~\ref{fig:figure_qualitative_davis} presents a qualitative comparison of the dynamic novel view synthesis performance of our MoBluRF against T-NeRF~\cite{gao2022monocular}, BAD-NeRF~\cite{wang2023bad}, and DyBluRF~\cite{sun2024dyblurf} on the DAVIS dataset~\cite{pont20172017}. As shown in Fig.~\ref{fig:figure_qualitative_davis}, our MoBluRF produces sharper novel view synthesis results than SOTA methods~\cite{gao2022monocular, wang2023bad, sun2024dyblurf}, effectively handling both static backgrounds and moving object regions in real-world blurry monocular video frames. However, BAD-NeRF~\cite{wang2023bad} struggles to reconstruct sharp moving objects due to its limited capacity to model scene dynamics. T-NeRF~\cite{gao2022monocular} exhibits blurriness and fails to preserve sharp details, such as the edges and patterns on helmets of moving objects. DyBluRF~\cite{sun2024dyblurf} generates rendering results with noisy artifacts and misalignments caused by its unstable camera pose optimization approach. Unlike these SOTA methods, our MoBluRF successfully preserves sharp details from blurry training frames in both static and dynamic regions. These results validate our hypothesis about motion blur caused by object deformations, demonstrating its applicability in practical settings.}

\begin{table}[h]
\centering
\caption{Robustness of MoBluRF against degrees of blurriness. `Trained w/ Original' refers to training each model with the original iPhone dataset~\cite{gao2022monocular}, while `Trained w/ Blurry' applies to the Blurry iPhone dataset.}
\vspace{-2mm}
\setlength\tabcolsep{4pt} 
\renewcommand{\arraystretch}{1.2}
\scalebox{0.9}{
\begin{tabular}{c|c|cccc}
\bottomrule
\hline\noalign{\smallskip}
Methods & Trained w/ & mPSNR$\uparrow$ & mSSIM$\uparrow$ & mLPIPS$\downarrow$ & tOF$\downarrow$ \\  
\bottomrule
\hline\noalign{\smallskip}
DyBluRF~\cite{sun2024dyblurf}       & Original
& \begin{tabular}[c]{c} 10.40\end{tabular}
& \begin{tabular}[c]{c} 0.444\end{tabular}
& \begin{tabular}[c]{c} 0.724\end{tabular}
& \begin{tabular}[c]{c} 2.585\end{tabular} \\
DyBluRF~\cite{sun2024dyblurf}       & Blurry
& \begin{tabular}[c]{c} 11.65 \vspace{-0.1cm} \\ \scriptsize \textcolor{lightorange}{\textbf{+1.25}} \end{tabular}
& \begin{tabular}[c]{c} 0.447 \vspace{-0.1cm} \\ \scriptsize \textcolor{lightorange}{\textbf{+0.003}} \end{tabular}
& \begin{tabular}[c]{c} 0.725 \vspace{-0.1cm} \\ \scriptsize \textcolor{lightorange}{\textbf{+0.001}} \end{tabular}
& \begin{tabular}[c]{c} 2.454 \vspace{-0.1cm} \\ \scriptsize \textcolor{lightorange}{\textbf{-0.131}} \end{tabular} \\
\midrule
T-NeRF~\cite{gao2022monocular}       & Original
& \begin{tabular}[c]{c} 16.96\end{tabular}
& \begin{tabular}[c]{c} 0.577\end{tabular}
& \begin{tabular}[c]{c} 0.379\end{tabular}
& \begin{tabular}[c]{c} 0.843\end{tabular} \\
T-NeRF~\cite{gao2022monocular}       & Blurry
& \begin{tabular}[c]{c} 16.15 \vspace{-0.1cm} \\ \scriptsize \textcolor{lightorange}{\textbf{-0.81}} \end{tabular}
& \begin{tabular}[c]{c} 0.547 \vspace{-0.1cm} \\ \scriptsize \textcolor{lightorange}{\textbf{-0.030}} \end{tabular}
& \begin{tabular}[c]{c} 0.456 \vspace{-0.1cm} \\ \scriptsize \textcolor{lightorange}{\textbf{+0.077}} \end{tabular}
& \begin{tabular}[c]{c} 0.990 \vspace{-0.1cm} \\ \scriptsize \textcolor{lightorange}{\textbf{+0.147}} \end{tabular} \\
\midrule
HyperNeRF~\cite{park2021hypernerf}   & Original & 16.81 & 0.569 & \textcolor{red}{\textbf{0.332}} & 0.869 \\
HyperNeRF~\cite{park2021hypernerf}   & Blurry
& \begin{tabular}[c]{c} 15.30 \vspace{-0.1cm} \\ \scriptsize \textcolor{lightorange}{\textbf{-1.51}} \end{tabular}
& \begin{tabular}[c]{c} 0.535 \vspace{-0.1cm} \\ \scriptsize \textcolor{lightorange}{\textbf{-0.034}} \end{tabular}
& \begin{tabular}[c]{c} 0.447 \vspace{-0.1cm} \\ \scriptsize \textcolor{lightorange}{\textbf{+0.115}} \end{tabular}
& \begin{tabular}[c]{c} 1.085 \vspace{-0.1cm} \\ \scriptsize \textcolor{lightorange}{\textbf{+0.216}} \end{tabular} \\
\midrule
\textbf{MoBluRF (Ours)}              & Original & \textcolor{red}{\textbf{17.37}} & \textcolor{red}{\textbf{0.591}} & \textcolor{blue}{\underline{0.373}} & \textcolor{red}{\textbf{0.664}} \\
\textbf{MoBluRF (Ours)}              & Blurry
& \begin{tabular}[c]{c} \textcolor{blue}{\underline{17.33}} \vspace{-0.1cm} \\ \scriptsize \textcolor{lightorange}{\textbf{-0.04}} \end{tabular}
& \begin{tabular}[c]{c} \textcolor{red}{\textbf{0.591}} \vspace{-0.1cm} \\ \scriptsize \textcolor{lightorange}{\textbf{-0.000}} \end{tabular}
& \begin{tabular}[c]{c} 0.376 \vspace{-0.1cm} \\ \scriptsize \textcolor{lightorange}{\textbf{+0.003}} \end{tabular}
& \begin{tabular}[c]{c} \textcolor{blue}{\underline{0.675}} \vspace{-0.1cm} \\ \scriptsize \textcolor{lightorange}{\textbf{+0.011}} \end{tabular} \\
\bottomrule
\hline\noalign{\smallskip}
\end{tabular}
}
\label{table:quantitative_robust}
\end{table}

\subsubsection{Robustness of MoBluRF for Different Degrees of Blurriness}
Table \ref{table:quantitative_robust} shows the robustness of MoBluRF for different degrees of blurriness by comparing with the other dynamic novel view synthes methods such as T-NeRF~\cite{gao2022monocular}, HyperNeRF~\cite{park2021hypernerf}, and DyBluRF~\cite{sun2024dyblurf}. Each method is trained with two datasets: the original iPhone dataset~\cite{gao2022monocular} or our newly synthesized Blurry iPhone dataset. As shown, both T-NeRF~\cite{gao2022monocular} and HyperNeRF~\cite{park2021hypernerf} experience substantial performance declines across all metrics with the Blurry iPhone dataset compared to their performances with the original iPhone dataset~\cite{gao2022monocular}, underscoring the challenge posed by our Blurry iPhone dataset for dynamic novel view synthesis without a deblurring module. \jm{On the other hand, the overall performance of DyBluRF~\cite{sun2024dyblurf}, trained on both datasets, is considerably lower than the other methods, and no meaningful tendency can be observed consistently between the two datasets.} In contrast, our MoBluRF maintains consistent results between the two datasets across all metrics, demonstrating its robustness to different degrees of blurriness.

\begin{figure*}
\centering
\includegraphics[scale=0.233]{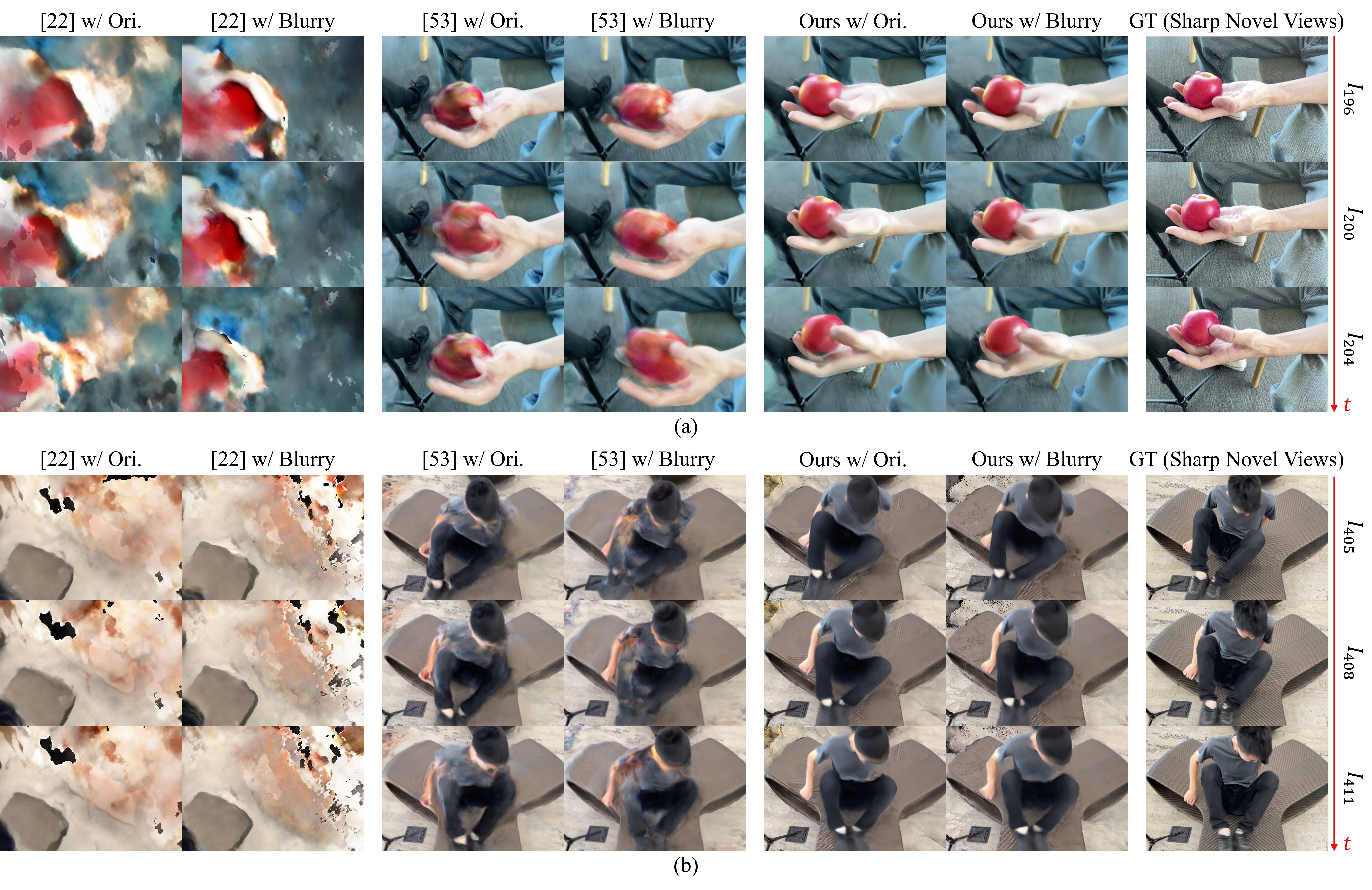}
\vspace{-0.5cm}
\caption{Novel view synthesis results of DyBluRF~\cite{sun2024dyblurf}, HyperNeRF~\cite{park2021hypernerf} and our MoBluRF when trained with different degrees of blurriness in (a) \textit{Apple} scene and (b) \textit{Spin} scene. We compare our MoBluRF, trained on either the original (`Ours w/ Ori.') or the Blurry (`Ours w/ Blurry') iPhone datasets, with DyBluRF~\cite{sun2024dyblurf} and HyperNeRF~\cite{park2021hypernerf}, also trained on either the original (`\cite{sun2024dyblurf} w/ Ori.' and `\cite{park2021hypernerf} w/ Ori.') or the Blurry (`\cite{sun2024dyblurf} w/ Blurry' and `\cite{park2021hypernerf} w/ Blurry') iPhone datasets.}
\label{fig:figure_qualitative_robust}
\end{figure*}

Fig. \ref{fig:figure_qualitative_robust} shows qualitative results comparing our MoBluRF with HyperNeRF~\cite{park2021hypernerf} and DyBluRF~\cite{sun2024dyblurf} when trained with the original iPhone dataset~\cite{gao2022monocular} or our Blurry iPhone dataset. We observe that HyperNeRF~\cite{park2021hypernerf} exhibits inconsistent artifacts in dynamic regions, \jm{while the rendered novel views of DyBluRF~\cite{sun2024dyblurf} are significantly misaligned across both datasets.} In contrast, MoBluRF's superior performance is attributed to its innovatively designed framework, which incorporates our novel $\mathcal{L}_{sm}$ and $\mathcal{L}_{lg}$ losses. These components are specifically tailored to optimize radiance fields effectively from highly dynamic monocular videos. Moreover, our MoBluRF exhibits consistent dynamic novel view synthesis performances when trained on the Blurry iPhone dataset, thanks to our effective motion-aware deblurring method which comprises the BRI and MDD stages for sharp renderings of dynamic novel views.
We also visually analyze the robustness of our MoBluRF for different degrees of blurriness in Fig. \ref{fig:figure_robust}. In Fig. \ref{fig:figure_robust}-(a), we show the rendered color $\hat{\bm{C}}^{full}$ of the base ray (green boxes) and the averaged rendered color $\hat{\bm{C}}_\mathcal{B}^{full}$ computed by Eq. \ref{eq:avg_color} when trained with the Blurry iPhone dataset or the original iPhone dataset~\cite{gao2022monocular}, respectively (blue boxes). As shown, our MoBluRF adaptively predicts latent sharp rays with the help of stably modeling the physical blur process, so maintaining the sharpness of the base ray's rendering color $\hat{\bm{C}}^{full}$ for various degrees of blurriness in the training frames. However, T-NeRF~\cite{gao2022monocular} and HyperNeRF~\cite{park2021hypernerf} produce blurrier results for the Blurry iPhone dataset compared to the original iPhone dataset~\cite{gao2022monocular}, as shown in Fig. \ref{fig:figure_robust}-(b). \jm{Besides, DyBluRF~\cite{sun2024dyblurf} fails to reconstruct reasonable geometry across both datasets. Furthermore, as shown in Fig. \ref{fig:figure_robust}-(c), the SOTA deblurring NeRFs~\cite{wang2023bad, lee2023dp, sun2024dyblurf} produce suboptimal deblurring reconstructions from blurry training frames, whereas our MoBluRF effectively addresses blurriness (green boxes).} We further demonstrate our MoBluRF's optimization capacity for adapting to varying degrees of blurriness in Fig. \ref{fig:figure_robust_extended}.
This figure shows the full rendered colors $\{\hat{\bm{C}}^{full}(\dot{\bm{r}}_{\bm{p};t;q})\}_{q=1}^{N_b}$ of all $N_b(=6)$ latent sharp rays, $\hat{\bm{C}}^{full}(\hat{\bm{r}}_{\bm{p};t})$ of the base ray, and the averaged rendered color $\hat{\bm{C}}_\mathcal{B}^{full}(\hat{\bm{r}}_{\bm{p};t})$ computed by Eq. \ref{eq:avg_color}, when our MoBluRF is trained with the Blurry iPhone dataset or the original iPhone dataset~\cite{gao2022monocular}.
As shown in Fig. \ref{fig:figure_robust_extended}-(a), our MoBluRF adaptively predicts diverse latent sharp rays based on the base ray to effectively render the averaged blurry color corresponding to the physical blur process (`GT (Train)') when trained with the Blurry iPhone dataset. On the other hand, in Fig. \ref{fig:figure_robust_extended}-(b), the rendered colors exhibit remarkable consistency across all latent sharp rays as well as the base ray. This consistency results in a relatively sharp averaged color that closely aligns with the training frames found in the original iPhone dataset~\cite{gao2022monocular}. We also compute the average standard deviations of the color (`$\sigma^{avg}_{C}$'), ray origin (`$\sigma^{avg}_{\mathbf{o}}$'), and ray direction (`$\sigma^{avg}_{\mathbf{d}}$'), respectively, for all latent sharp rays and the base ray, as trained with each dataset. As demonstrated, the standard deviations of all values are significantly larger when trained with the Blurry iPhone dataset than with the original iPhone dataset~\cite{gao2022monocular}. This underscores the robustness of our MoBluRF's optimization capacity for various degrees of blurriness.

\begin{figure*}
\centering
\includegraphics[width=\textwidth]{Figures/robustness_degree.jpg}
\vspace{-3mm}
\caption{Robustness of MoBluRF for different degrees of blurriness. (a) `$\hat{\bm{C}}^{full}$' and `$\hat{\bm{C}}_\mathcal{B}^{full}$' refers to the rendering of the base rays and the averaged rendering computed with Eq. \ref{eq:avg_color}, respectively, by our MoBluRF. (b) `w/ Blurry' and `w/ Ori.' refer to the synthesized novel views by each method when trained with the Blurry iPhone dataset or the original iPhone dataset~\cite{gao2022monocular}, respectively. (c) Comparisons given the original \cite{gao2022monocular}'s poses. `Ours ($\hat{\bm{C}}^{full}$)' refers to the sharp rendering by our MoBluRF trained with blurry frames (`Blurry'), which appears similar to the corresponding original training frames (`Original'), whereas other methods~\cite{gao2022monocular, wang2023bad, lee2023dp, sun2024dyblurf} fail to learn the sharp radiance fields from the blurry frames.}
\vspace{-3mm}
\label{fig:figure_robust}
\end{figure*}

\begin{figure*}
\centering
\includegraphics[scale=0.17]{Figures/figure_adaptive.jpg}
\vspace{-3mm}
\caption{Adaptive prediction of latent sharp colors against different degrees of blurriness. We visualize the rendering frames which refer to each rendered color $\hat{\bm{C}}^{full}(\dot{\bm{r}}_{\bm{p};t;q})$ of $q^{\text{th}}$ latent sharp ray, $\hat{\bm{C}}^{full}(\hat{\bm{r}}_{\bm{p};t})$ of the base ray, and the averaged rendered color $\hat{\bm{C}}_\mathcal{B}^{full}(\hat{\bm{r}}_{\bm{p};t})$ computed with Eq. \ref{eq:avg_color}. These are produced by our MoBluRF when trained with (a) the Blurry iPhone dataset and (b) the original iPhone dataset~\cite{gao2022monocular}, respectively. `$\sigma^{avg}_{C}$', `$\sigma^{avg}_{\mathbf{o}}$', and `$\sigma^{avg}_{\mathbf{d}}$' refer to the average standard deviations of the color, ray origin, and ray direction, respectively, for all latent sharp rays and the base ray.}
\vspace{-3mm}
\label{fig:figure_robust_extended}
\end{figure*}

\vspace{-2mm}
\subsection{Ablation Study on Blurry iPhone dataset}
We conduct an ablation study to analyze the effectiveness of our $\mathcal{L}_{sm}$ and $\mathcal{L}_{lg}$ for dynamic scene reconstruction, and our two-stage training strategy of the BRI and MDD stages for deblurring dynamic radiance fields. Table \ref{table:ablation_study} presents detailed quantitative results for the average performance across all scenes.

\begin{table}
\centering
\caption{Ablation study conducted on the Blurry iPhone dataset. `$\mathcal{L}_\mathcal{D}$' is \cite{li2021neural}'s depth loss. `$\mathcal{N}$' is the naive pose optimization~\cite{nerfmm}, i.e., without interleaved optimization. `2D' is preprocessing with GShiftNet~\cite{li2023simple}. \jm{‘Bent’ refers to the variant which bends the latent sharp rays, instead of refining the straight rays.}}
\vspace{-2mm}
\centering
\setlength\tabcolsep{2.8pt} 
\renewcommand{\arraystretch}{1.3}
\scalebox{0.85}{
\begin{tabular}{ c| c c c c | c c c c }
\bottomrule
\hline\noalign{\smallskip}
Variants & $\mathcal{L}_{sm}$ & $\mathcal{L}_{lg}$ & BRI & MDD & mPSNR$\uparrow$ & mSSIM$\uparrow$ & mLPIPS$\downarrow$ & tOF$\downarrow$ \\  
\bottomrule
\hline\noalign{\smallskip}
(a) & - & - & - & -                                    & 16.48 & 0.563 & 0.455 & 0.883 \\
(b) & $\checkmark$ & - & - & -                         & 16.85 & 0.572 & 0.422 & 0.714 \\
(c) & $\checkmark$ & $\mathcal{L}_\mathcal{D}$ & -  &       -      &  15.13 & 0.528 & 0.666 & 1.063 \\
(d) & $\checkmark$ & $\checkmark$  &     - &     -     & 17.18 & 0.590 & 0.426 & 0.692 \\
(e) & $\checkmark$ & $\checkmark$  &     - &     2D     & 16.80 & 0.582 & 0.413 & 0.740 \\
(f) & $\checkmark$  & $\checkmark$ & $\mathcal{N}$ &       -      & 16.06 & 0.545 & 0.449 & 0.806 \\
(g) & $\checkmark$ & $\checkmark$ & -  & $\checkmark$ & 16.86 & 0.568 & 0.434 & 0.724 \\
(h) & $\checkmark$ & $\checkmark$  & $\checkmark$  &       -      & 17.24 & \textcolor{red}{\textbf{0.593}} & 0.413 & \textcolor{blue}{\underline{0.653}}     \\
(i) & $\checkmark$  & $\checkmark$ & $\checkmark$  & Bent & \textcolor{blue}{\underline{17.30}} & 0.533 & \textcolor{blue}{\underline{0.380}} & \textcolor{red}{\textbf{0.649}} \\
(j) & $\checkmark$ & $\checkmark$  & $\checkmark$  & w/o GMRP & 17.28 & 0.584 & 0.403 & 0.692 \\
(k) & $\checkmark$ & $\checkmark$  & $\checkmark$  & $\checkmark$ & \textcolor{red}{\textbf{17.33}} & \textcolor{blue}{\underline{0.591}} & \textcolor{red}{\textbf{0.376}} & 0.675 \\
\bottomrule
\hline\noalign{\smallskip}
\end{tabular}}
\label{table:ablation_study}
\end{table}

\begin{figure}
\centering
\includegraphics[width=0.49\textwidth]{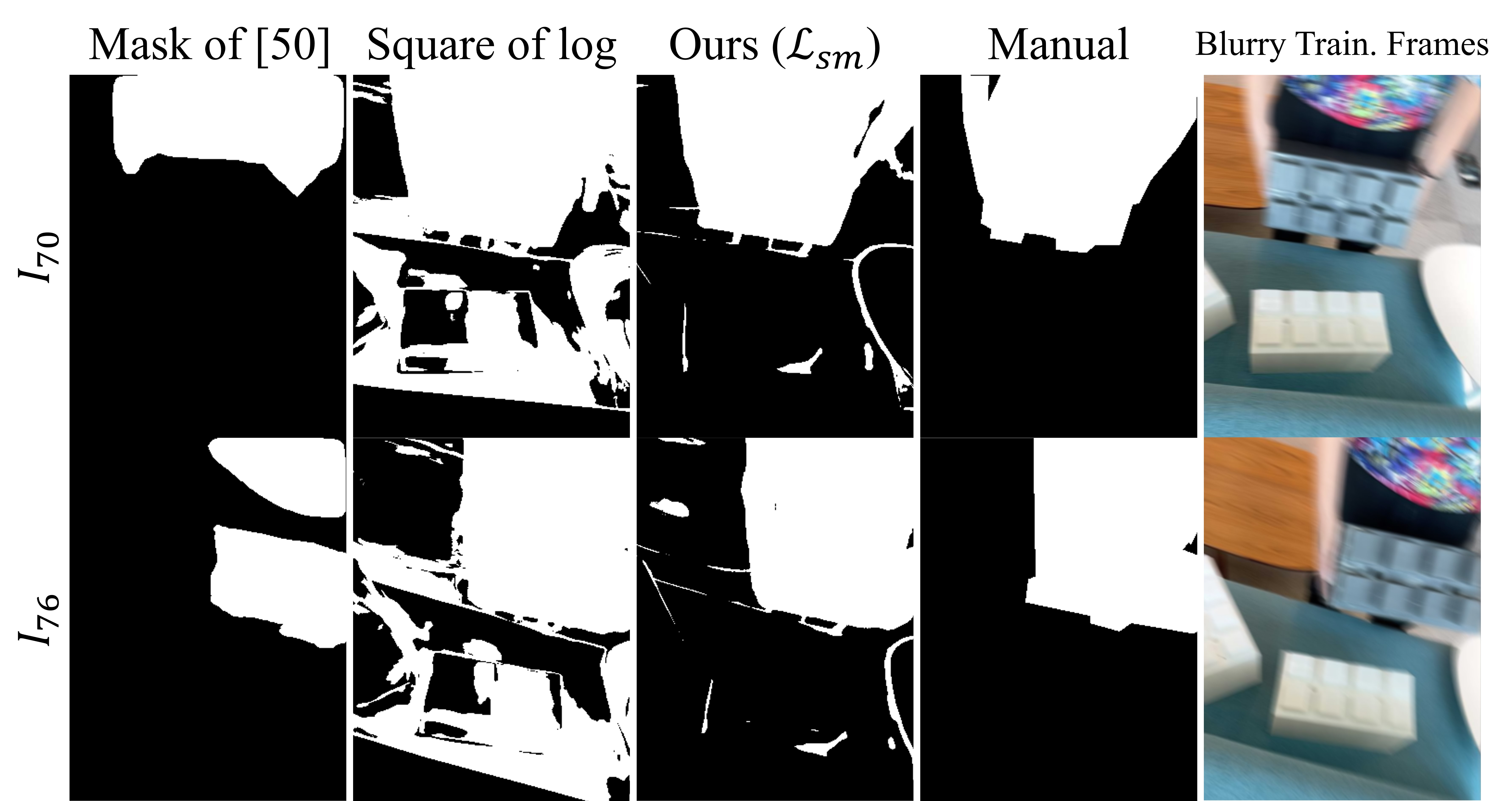}
\vspace{-6mm}
\caption{Ablation study on $\mathcal{L}_{sm}$. `Mask of \cite{liu2023robust}' indicates the preprocessed motion mask using optical flows~\cite{teed2020raft} from only consecutive frames and `Square of log' indicates the variant of our $\mathcal{L}_{sm}$ to the squared value of the logarithm. `Manual' indicates the manual annotation mask from Roboflow$^{TM}$ \cite{roboflow}. `Blurry Train. Frames' indicates the corresponding blurry train frames of our Blurry iPhone dataset. Note that our motion mask is all white when trained without $\mathcal{L}_{sm}$.}
\label{fig:ablation_sm_loss}
\end{figure}

\noindent \textbf{Unsupervised Staticness Maximization Loss ($\mathcal{L}_{sm}$).}
By incorporating $\mathcal{L}_{sm}$, MoBluRF effectively decomposes 3D scenes into static and dynamic components in an \textit{unsupervised} manner. This aids our framework in more robustly reconstructing dynamic scenes, leading to performance gains as shown in variant (b) compared to variant (a) in Table \ref{table:ablation_study}. It is also worth noting that previous methods \cite{liu2023robust, gao2021dynamic, li2021neural} only consider optical flows \cite{teed2020raft} for temporally close time instances to generate motion masks. This approach makes capturing long-range temporal cues challenging and yields inconsistent motion masks, as shown in Fig. \ref{fig:ablation_sm_loss}. These unstable masks can lead to the misclassification of dynamic objects as static regions, undermining the multi-view consistency of the Static Net $F_{\theta^s}$. In contrast, MoBluRF produces more consistent masks along the whole temporal axis, thanks to our $\mathcal{L}_{sm}$ for unsupervised $\bm{M}(\hat{\bm{r}}_{\bm{p};t})$ training. Additionally, Fig. \ref{fig:ablation_sm_loss} shows that the absolute value of the logarithm in $\mathcal{L}_{sm}$ helps overcome noisy motion prediction in $\bm{M}(\hat{\bm{r}}_{\bm{p};t})$, compared to the squared value of the logarithm (`Square of log').

\begin{figure}
\centering
\includegraphics[width=0.49\textwidth]{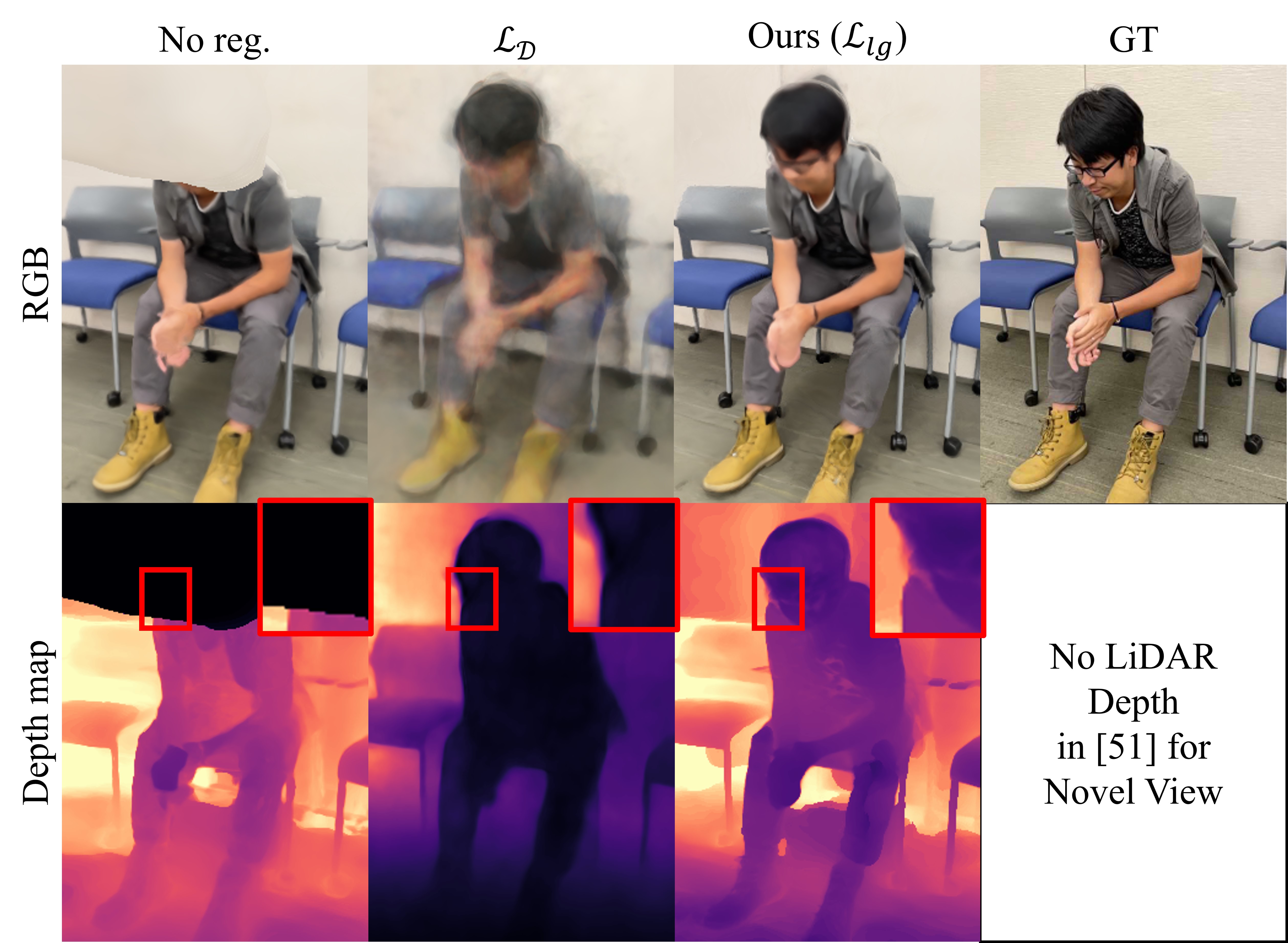}
\vspace{-6mm}
\caption{Ablation study on $\mathcal{L}_{lg}$. `No reg.' indicates the variant without any loss for geometry. `Ours' refers to our $\mathcal{L}_{lg}$.}
\label{fig:ablation_lg_loss}
\end{figure}

\noindent \textbf{Local Geometry Variance Distillation ($\mathcal{L}_{lg}$).}
In Table \ref{table:ablation_study}, our $\mathcal{L}_{lg}$ in variant (d) significantly improves the accuracy of 3D structures, surpassing variant (b) which does not incorporate geometry regularization. This enhancement is substantiated by higher mPSNR and mSSIM scores, and more consistent deformation across dynamic scenes, as denoted by the improved tOF. Additionally, variant (d) showcases the robustness of our $\mathcal{L}_{lg}$ that is more effective than the scale-invariant depth loss (`$\mathcal{L}_\mathcal{D}$') \cite{li2021neural} in variant (c).
Fig. \ref{fig:ablation_lg_loss} visualizes the effectiveness of our $\mathcal{L}_{lg}$ by comparing it to the variants without geometry regularization (`No reg.') and with the scale-invariant depth loss~\cite{li2021neural}. Our $\mathcal{L}_{lg}$ leads to better reconstruction of geometry for dynamic objects and novel view synthesis than the others.

\begin{figure}
\centering
\includegraphics[width=0.49\textwidth]{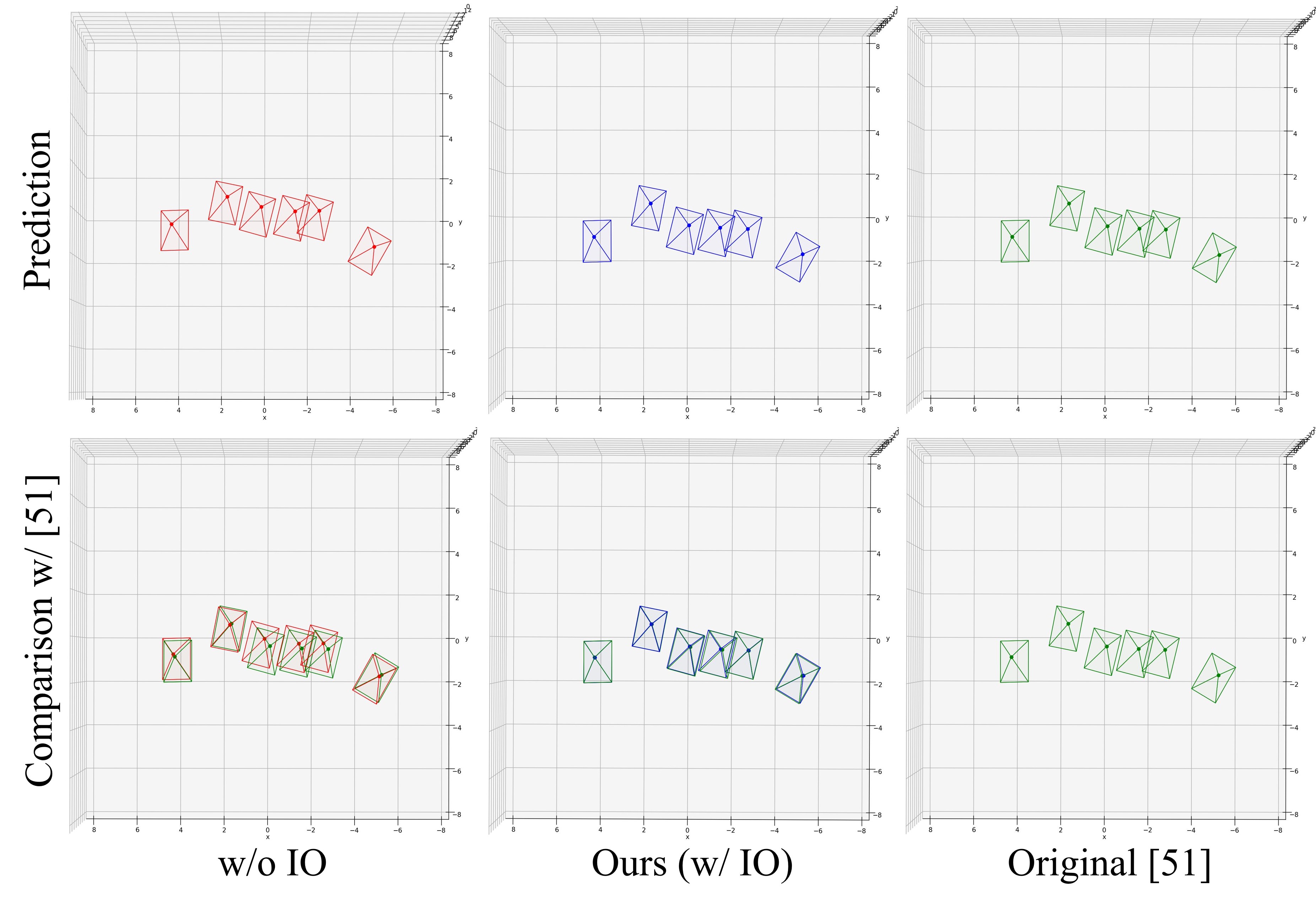}
\vspace{-6mm}
\caption{\jm{Ablation study on interleaved optimization. ‘w/o IO’ and ‘Ours (w/ IO)’ refer to the refined camera poses after BRI stage without and with interleaved optimization, respectively. These refined camera poses are used to compute the base rays. ‘Original~\cite{gao2022monocular}’ refers to the corresponding camera poses of the original iPhone dataset~\cite{gao2022monocular}.}}
\label{fig:ablation_io}
\end{figure}

\noindent \textbf{BRI Stage.}
Our BRI in variant (h) outperforms variant (d) across all metrics. We analyze the critical role of our interleaved optimization strategy in the BRI stage by omitting it in variant (f). As shown in Table \ref{table:ablation_study}, refining the base ray while optimizing radiance fields without the interleaved optimization leads to adverse effects. As discussed in Sec. \ref{sect:design_consideration}, $F_{\theta^d}$ can inappropriately compensate for the base ray inaccuracies obtained by learning unnecessary deformations. Hence, incorporating static geometry cues for precise base ray initialization is crucial, facilitated by our interleaved optimization. Moreover, applying the MDD stage directly without a well-initialized base ray can deteriorate performance, as evidenced by the substantial performance gap between \quan{variants (k)} and (g). This indicates that proper initialization of base rays by the BRI stage is crucial for subsequent deblurring. \jm{Fig.~\ref{fig:ablation_io} and Table~\ref{table:ablation_io} demonstrate the effectiveness of our proposed interleaved optimization strategy during the BRI stage. Our full model with interleaved optimization (`Ours (w/ IO)') predicts more accurate refined camera poses that are closer to the original iPhone dataset~\cite{gao2022monocular} compared to the ablated model without interleaved optimization (`w/o IO').}

\begin{table}[h]
\small
\caption{\jm{Ablation study on interleaved optimization conducted on the Blurry iPhone dataset. ‘w/o IO’ and ‘Ours (w/ IO)’ refer to the refined camera poses after BRI stage without and with interleaved optimization, respectively. These refined camera poses are used to compute the base rays. Error$_{rot}$ ($^{\circ}$)$\downarrow$ and Error$_{trans}$$\downarrow$ refer to the average rotation and translation error between the camera poses of each model and the camera poses of the original iPhone dataset~\cite{gao2022monocular}, respectively.
}}
\begin{center}
\setlength\tabcolsep{4pt} 
\renewcommand{\arraystretch}{1.3}
\scalebox{0.85}{
\begin{tabular}{c | c c}
\bottomrule
\hline\noalign{\smallskip}
Variants & Error$_{rot}$ ($^{\circ}$)$\downarrow$ & Error$_{trans}$$\downarrow$ \\  
\bottomrule
\hline\noalign{\smallskip}
w/o IO  & 0.00914 & 0.00899 \\
Ours (w/ IO) & \textbf{0.00871} & \textbf{0.00697}  \\
\bottomrule
\hline\noalign{\smallskip}
\end{tabular}}
\end{center}
\label{table:ablation_io}
\vspace{-0.2cm}
\end{table}

\begin{figure}
\centering
\includegraphics[width=0.49\textwidth]{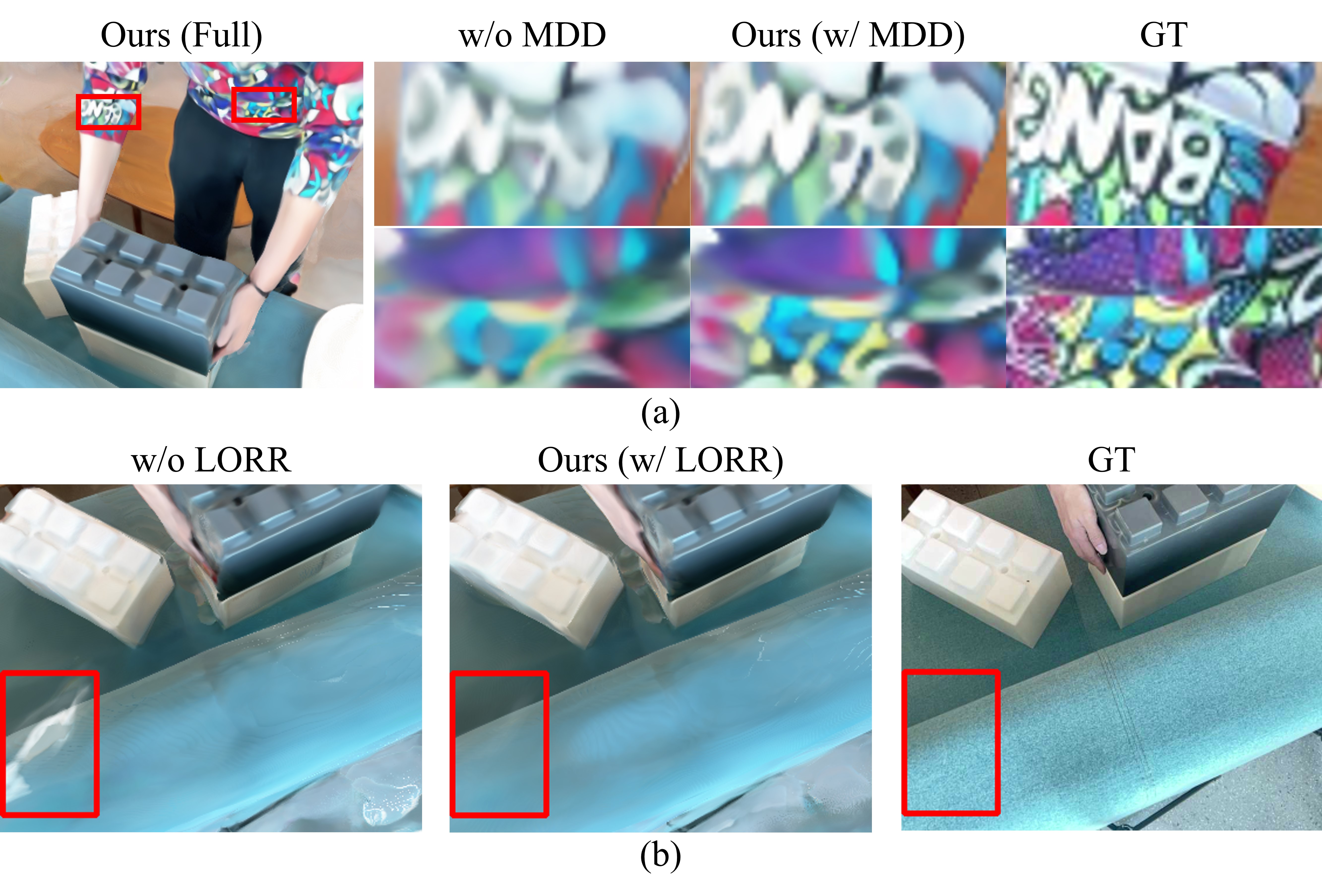}
\vspace{-8mm}
\caption{Ablation study on (a) the MDD stage (Sec. \ref{sect:deblur_module}) and (b) LORR (Eq. \ref{eq:local_motion_ray}) conducted on the Blurry iPhone dataset.}
\label{fig:ablation_mdd}
\end{figure}

\begin{figure}
\centering
\includegraphics[width=0.49\textwidth]{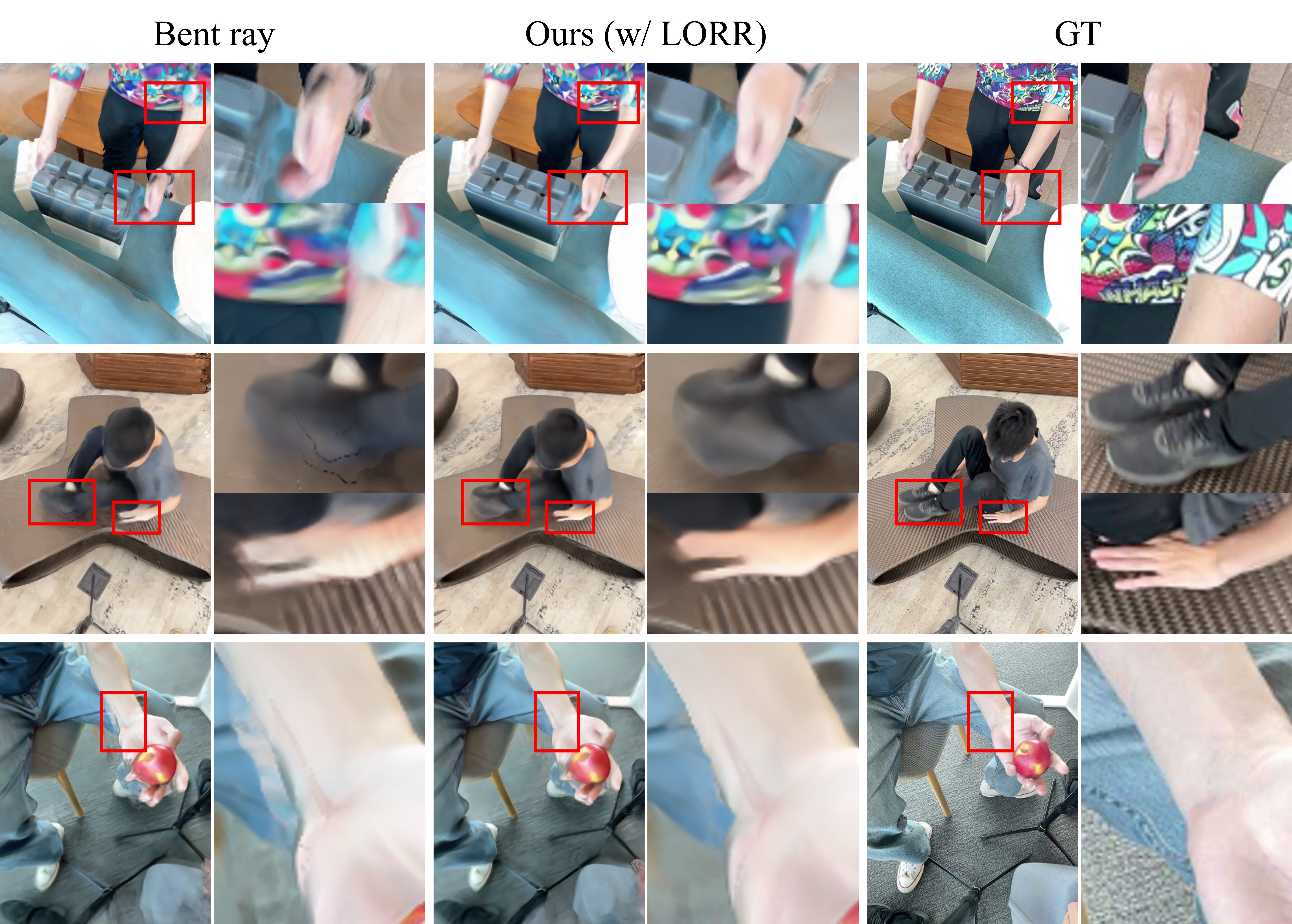}
\vspace{-5mm}
\caption{\jm{Ablation study on LORR (Eq. \ref{eq:local_motion_ray}) conducted on the Blurry iPhone dataset. `Bent ray' refers to the variant which bends the latent sharp rays, instead of refining the straight rays. In this ablated model, $F_{\theta^l}$ predicts the offsets for \textit{all sampling points} along the latent sharp rays. Then, the offsets are added to the corresponding sampling points.}}
\label{fig:ablation_lorr}
\end{figure}

\noindent \textbf{MDD Stage.}
In Table \ref{table:ablation_study}, by comparing the final MoBluRF \quan{(k)} with variant (h), we observe that the MDD stage significantly enhances perceptual quality, as evidenced by the improved mLPIPS scores. \quan{Fig. \ref{fig:ablation_mdd}-(a) illustrates the effectiveness of our MoBluRF in novel view synthesis for video deblurring, demonstrating the MDD stage's ability to adeptly handle blurriness in both static and dynamic regions.} As shown in Fig. \ref{fig:ablation_mdd}-(b), our MoBluRF more delicately decomposes the mixture of camera and object motions (i.e., white brick) as the training proceeds, leading to robust novel view synthesis for a region where the object passes during training, when compared to our variant trained without LORR (Eq. \ref{eq:local_motion_ray}). \jm{To further validate the effectiveness of our LORR as an additional regularization prior, which refines the straight rays in dynamic regions, we compare the final MoBluRF \quan{(k)} with variant (i) that bends the rays. In this ablated model, the local object-motion MLP $F_{\theta^l}$ predicts the offsets for all sampling points along the latent sharp rays, replacing the screw-axis-based ray warping mechanism. These offsets are subsequently applied to the corresponding sampling points, resulting in bent rays (`Bent'). The experimental results demonstrate that our LORR consistently outperforms variant (i) across all rendering quality metrics (mPSNR, mSSIM, mLPIPS), particularly excelling in structural similarity (mSSIM). This highlights LORR's ability to provide a more effective and robust prior for the regularization of radiance fields. Variant (i) achieves better tOF scores than our full model. However, this is primarily due to the static background regions that are not observed in the training frames and are excluded using the co-visibility mask~\cite{gao2022monocular} when measuring the other three rendering quality metrics. Fig.~\ref{fig:ablation_lorr} illustrates visual comparisons between our final MoBluRF with LORR and the ablated model that employs ray bending for latent sharp rays. As shown in Fig.~\ref{fig:ablation_lorr}, our final model consistently reconstructs sharper and more distinct boundaries for moving objects. In contrast, the ablated model struggles with unnatural repetitive artifacts in dynamic regions, particularly around moving objects, highlighting the robustness of our LORR in preserving spatial fidelity.} \quan{We also validate the necessity of GMRP in variant (j). As shown, the deblurring effect of the MDD stage weakens without GMRP, resulting in a worse LPIPS score compared to variant (k). The analysis of variants (i) and (j) emphasizes the effectiveness of our design choices for the MDD stage.} Our final MoBluRF \quan{(k)} significantly outperforms variant (e) and the cascade of GShiftNet~\cite{li2023simple} and variant (d), which evidences the advantages of full MoBluRF. Furthermore, variant (e) exhibits a slight improvement in mLPIPS but yields lower mPSNR, mSSIM, and tOF compared to variant (d). This suggests that even though the deblurring preprocess in the cascade approaches reduces blurriness in pixel domain, it is hard to learn precise spatio-temporal radiance fields due to the artifacts remained in the deblurred frames.

\noindent \textbf{The Number of Latent Sharp Rays.}
We provide a quantitative comparison of our MoBluRF model with varying numbers of latent sharp rays for the MDD stage in Table \ref{table:ablation_number_latent}. In particular, increasing the number $N_b$ of latent sharp rays can improve the reconstruction performance but it requires more computational cost. \jm{For mLPIPS, MoBluRF consistently demonstrates improved results with an increasing number of latent sharp rays, striking an optimal balance between rendering performance and training efficiency when using $N_b=6$. However, when $N_b$ is set to a large value, such as 10, the performance declines. This drop is likely attributed to optimization instability caused by the overfitting tendencies of an excessively large model.} To measure the training time per iteration, we utilize a single RTX 3090Ti GPU and set the batch size to 128.
It is notable that the $N_b$ only affects the training time. In the inference where we render a single predicted sharp ray per pixel, the inference time remains constant across all variants.

\begin{table}[h]
\vspace{-2mm}
\caption{\jm{Ablation Study on the number of latent sharp rays conducted on Blurry iPhone dataset.}}
\vspace{-5mm}
\begin{center}
\setlength\tabcolsep{4pt} 
\renewcommand{\arraystretch}{1.3}
\scalebox{0.85}{
\begin{tabular}{  c | c c c c | c}
\bottomrule
\hline\noalign{\smallskip}
$N_b$  & mPSNR$\uparrow$ & mSSIM$\uparrow$ & mLPIPS$\downarrow$ & tOF$\downarrow$ & Training time per iter. (s)\\  
\bottomrule
\hline\noalign{\smallskip}
2  & 16.90 & 0.571 & 0.409 & 0.750 & 0.095\\
4  & 17.37 & 0.592 & 0.387 & 0.668 & 0.137 \\
6  & 17.33 & 0.591 & 0.376 & 0.675 & 0.161\\
8  & 17.25 & 0.589 & 0.376 & 0.655 & 0.196 \\
10 & 17.19 & 0.579 & 0.383 & 0.675 & 0.226\\
\bottomrule
\hline\noalign{\smallskip}
\end{tabular}}
\end{center}
\label{table:ablation_number_latent}
\end{table}

\subsection{Ablation Study on Stereo Blur dataset}
\begin{figure}
\centering
\includegraphics[width=0.49\textwidth]{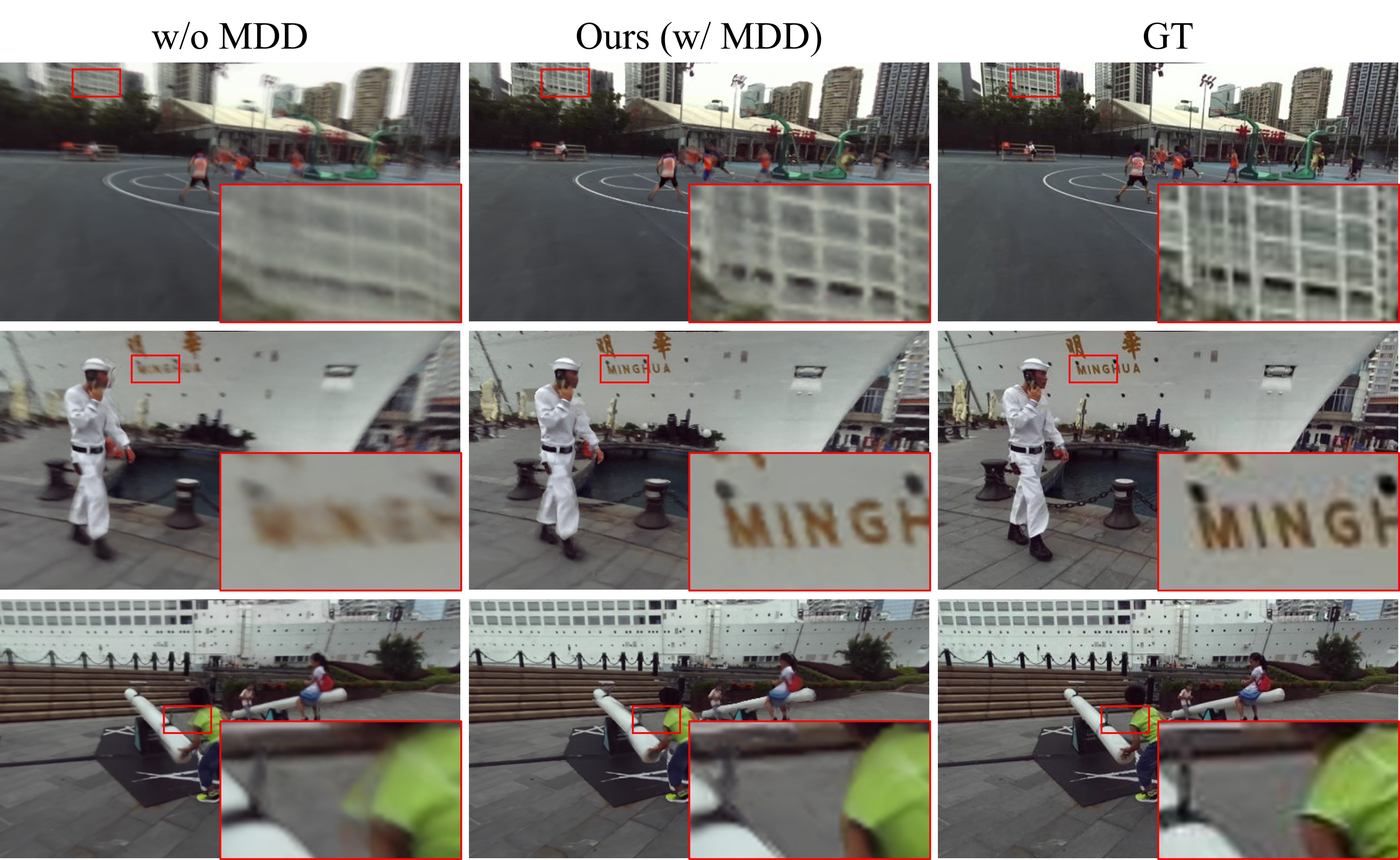}
\vspace{-6mm}
\caption{\jm{Ablation study on the MDD stage (Sec. \ref{sect:deblur_module}) conducted on the Stereo Blur dataset~\cite{zhou2019davanet}.}}
\label{fig:ablation_mdd_dyblurf}
\end{figure}

\quan{We further conduct an additional ablation study on the Stereo Blur dataset to evaluate the core components of MoBluRF. As shown in Table \ref{table:ablation_study_stereo}, the two regularization terms (Reg.), $\mathcal{L}_{sm}$ and $\mathcal{L}_{lg}$, significantly improve performance, as indicated by the enhancement in the `w/o BRI, w/o MDD' variant compared to the `Baseline (w/o Reg., w/o BRI, w/o MDD)' variant. However, without the deblurring components, the `w/o BRI, w/o MDD' variant fails to achieve optimal LPIPS and tOF scores. Similar to the trends observed in variant (g) of Table \ref{table:ablation_study}, the suboptimal performance of the `w/o BRI' variant underscores the critical importance of the BRI stage for the subsequent deblurring process in the MDD stage. On the other hand, the deblurring effectiveness of the MDD stage is demonstrated by the significant improvement in our final MoBluRF model, compared to the variants lacking the deblurring module, such as `w/o BRI, w/o MDD' and `w/o MDD'. Fig. \ref{fig:ablation_mdd_dyblurf} illustrates the corresponding qualitative comparisons for the MDD stage ablation. Additionally, consistent with variant (j) in Table \ref{table:ablation_study}, we investigate the ray warping design in the MDD stage. The decomposition into GMRP and LORR is important for the effectiveness of MDD, as evidenced by the deterioration in perceptual metrics in the 'w/o GMRP' variant, compared to the full MoBluRF model.}
\begin{table}
\centering
\caption{Ablation study conducted on the Stereo Blur dataset.}
\vspace{-2mm}
\centering
\setlength\tabcolsep{2.8pt} 
\renewcommand{\arraystretch}{1.3}
\scalebox{0.85}{
\begin{tabular}{ l | c c c c }
\bottomrule
\hline\noalign{\smallskip}
Variants & PSNR$\uparrow$ & SSIM$\uparrow$ & LPIPS$\downarrow$ & tOF$\downarrow$ \\  
\bottomrule
\hline\noalign{\smallskip}
Baseline (w/o Reg., w/o BRI, w/o MDD)                                    & 22.52 & 0.808 & 0.165 & 1.640 \\
w/o BRI, w/o MDD     & 25.52 & 0.883 & 0.165 & 1.639 \\
w/o BRI & 20.44 & 0.765 & 0.178 & 1.608 \\
w/o MDD    & 25.58 & 0.882 & 0.161 & 1.642     \\
w/o GMRP & \textcolor{red}{\textbf{25.78}} & \textcolor{blue}{\underline{0.887}} &\textcolor{blue}{\underline{0.145}} & \textcolor{blue}{\underline{1.421}} \\
MoBluRF (Ours)& \textcolor{blue}{\underline{25.69}} & \textcolor{red}{\textbf{0.893}} & \textcolor{red}{\textbf{0.078}} & \textcolor{red}{\textbf{0.816}} \\
\bottomrule
\hline\noalign{\smallskip}
\end{tabular}}
\label{table:ablation_study_stereo}
\end{table}

\subsection{Complexity Comparison}
We compare the inference time and the number of parameters of our MoBluRF to the existing methods by using a single RTX 3090 Ti in Table \ref{table:complexity}. Please note that our framework primarily focuses on improving the handling of motion blur problem in learning neural radiance fields, not on the rendering speed.

\begin{table}[h]
\centering
\setlength\tabcolsep{4pt} 
\renewcommand{\arraystretch}{1.3}
\vspace{-2mm}
\caption{Complexity comparison. We measure the performances on a single RTX 3090 Ti for the image resolution of $480\times360$.}
\vspace{-2mm}
\scalebox{1}{ 
\begin{tabular}{c | c c}
\bottomrule
\hline\noalign{\smallskip}
Methods & Inference Time (sec/frame) & \# of Parameters (M)  \\ \bottomrule
TiNeuVox~\cite{fang2022fast} & 1.5 & 24.83 \\
HexPlane~\cite{cao2023hexplane} & 2.2 & 9.73 \\
T-NeRF~\cite{gao2022monocular} & 5.0 & 1.19 \\
HyperNeRF~\cite{park2021hypernerf} & 7.5 & 1.31 \\ 
4D-GS~\cite{wu20234d} & 0.05 & 39.29 \\
DP-NeRF$_{t}$~\cite{lee2023dp} & 4.3 & 1.34 \\
BAD-NeRF$_{t}$~\cite{wang2023bad} & 4.4 & 1.22 \\ 
DyBluRF~\cite{sun2024dyblurf} & 3.7 & 1.21 \\ 
MoBluRF (Ours) & 9.1 & 1.78 \\
\bottomrule
\hline\noalign{\smallskip}
\end{tabular}
}
\label{table:complexity}
\end{table}

\subsection{Limitations and Future Work}
\begin{figure}[h]
    \centering
    \includegraphics[width=0.43\textwidth]{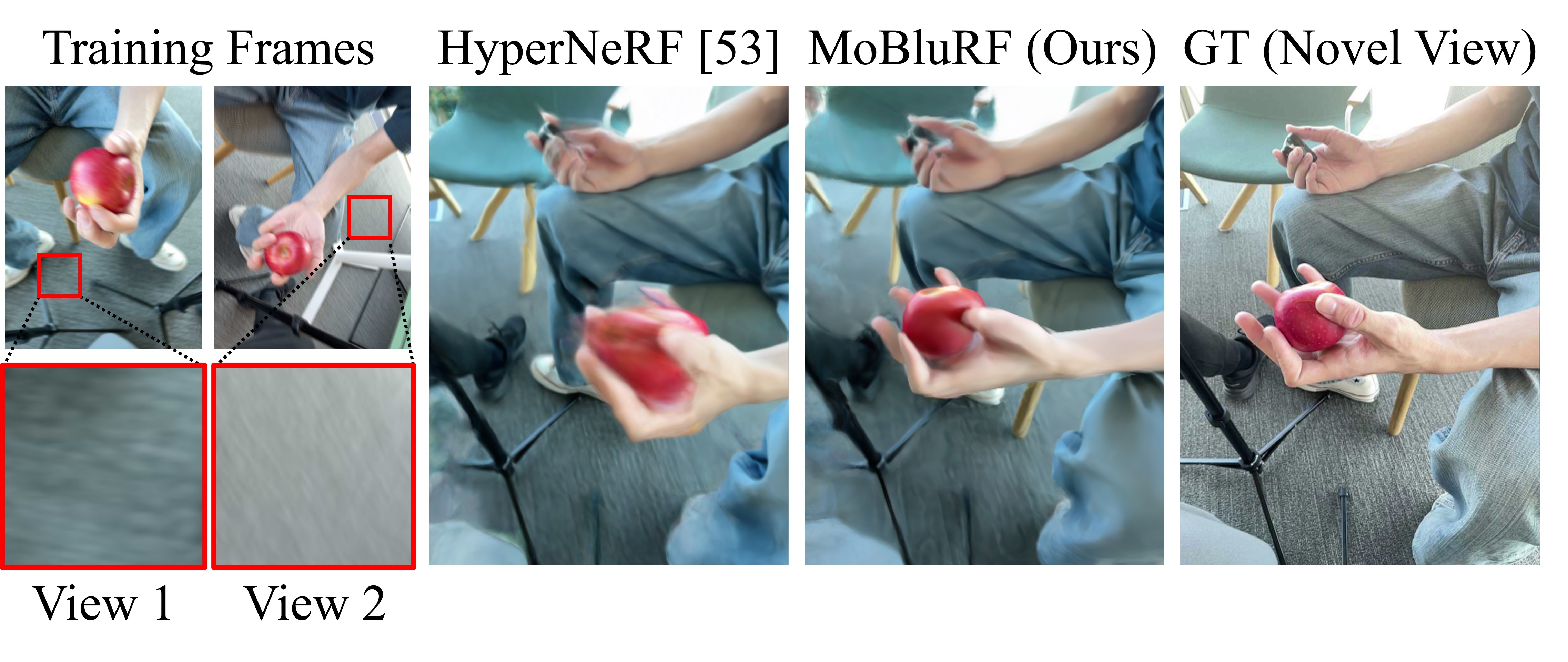}
    \vspace*{-5mm}
    \caption{Limitations of our MoBluRF. Since the majority of training video frames in the Blurry iPhone dataset exhibit a similar tone in View 1 (about $92\%$ of the training frames), the accurate reconstruction for validation views with a similar tone in View 2 (about $8\%$ of the training frames) becomes challenging. This difficulty arises due to the varying lighting effects present in the Blurry iPhone dataset.
    }
    \label{fig:limitation}
\end{figure}

In the Blurry iPhone dataset, there is one continuous training monocular video for each scene. Additionally, two fixed cameras with novel views are provided for evaluation, following the evaluation protocol of the original iPhone dataset~\cite{gao2022monocular}. This configuration makes it challenging to train deep-learning-based methods on unseen information that may be required for evaluation. For instance, in some specific scenes, as shown in Fig. \ref{fig:limitation}, due to different lighting environments, the tone of validation views can only be observed in a significantly small number of training samples (View 2). Hence, all kinds of radiance fields are generally overfitted to the major lighting condition of View 1. This challenge arises from the dataset's characteristics, which exhibit extremely diverse training and validation views. Relighting techniques that model the reflectance components or balancing training data can be one of options to handle this issue and we leave it as our future work.

On the other hand, a very recent work such as Gaussian Splatting-based dynamic methods, e.g., 4D-GS~\cite{wu20234d}, brings improved training and rendering efficiencies. However, according to our current experiments of Table \ref{table:quantitative_comparison} and the demo video, 4D-GS~\cite{wu20234d} struggles with handling highly monocular videos, as seen in our Blurry iPhone dataset and the original iPhone dataset~\cite{gao2022monocular}. Therefore, we plan to enhance our framework by integrating the Gaussian Splatting shading network into our effective motion-aware deblurring strategy as our future work.

\section{Conclusion}
We propose a novel motion deblurring NeRF for blurry monocular video, called MoBluRF, which can effectively render the sharp novel spatio-temporal views from blurry monocular frames. Our MoBluRF consists of a two-stage framework, including the BRI stage and the MDD stage. The BRI stage simultaneously reconstructs dynamic 3D scenes and initializes better base rays. The MDD stage introduces a novel ILSP approach to decompose latent sharp rays into global camera and local object motion components. We also incorporate simple yet effective $\mathcal{L}_{sm}$ and $\mathcal{L}_{lg}$ for stable training dynamic radiance field. Experimental results demonstrate that MoBluRF outperforms recent SOTA methods both qualitatively and quantitatively.

\section*{Acknowledgement}
This work was supported by Institute of Information and communications Technology Planning and Evaluation (IITP) grant funded by the Korean Government [Ministry of Science and ICT (Information and Communications Technology)] (Project Number: RS-2022-00144444, Project Title: Deep Learning Based Visual Representational Learning and Rendering of Static and Dynamic Scenes, 100\%).

\bibliographystyle{IEEEtran}
\bibliography{main_bbl.bib}

\vspace*{-1.3cm}
\begin{IEEEbiography}[{\includegraphics[width=1in,height=1.25in,clip,keepaspectratio]{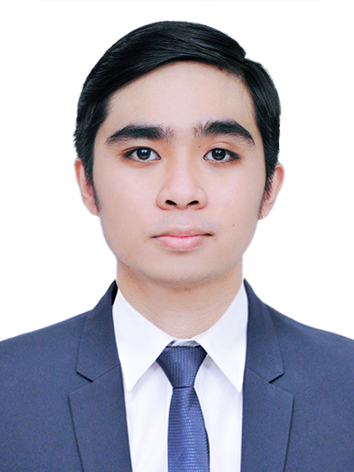}}]{Minh-Quan Viet Bui} received a B.S. degree in computer science from Ho Chi Minh City University of Technology (HCMUT), Ho Chi Minh City, Vietnam in 2021. He has been a Ph.D. student at the School of Electrical Engineering at Korea Advanced Institute of Science and Technology (KAIST), Daejeon, South Korea since 2022. His research interests include neural rendering, computer vision, and deep learning. He has published 3 papers related to computer vision in international journals, including PeerJ Computer Science and Image and Vision Computing.
\end{IEEEbiography}

\begin{IEEEbiography}[{\includegraphics[width=1in,height=1.25in,clip,keepaspectratio]{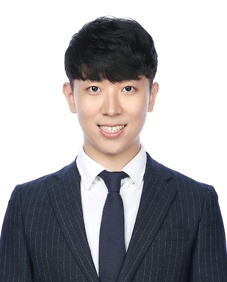}}]{Jongmin Park} received his B.E. and M.E. degrees in Electrical Engineering from KAIST, in 2020 and 2022, respectively. He has been a Ph.D. student at the School of Electrical Engineering at Korea Advanced Institute of Science and Technology (KAIST), Daejeon, South Korea since 2022. His research interests include novel view synthesis, Image/Video Restoration, and deep learning. He has published 1 paper at ICCV as the first author. His personal homepage can be found at https://sites.google.com/view/jongmin-park.
\end{IEEEbiography}

\vspace*{-5mm}
\begin{IEEEbiography}[{\includegraphics[width=1in,height=1.25in,clip,keepaspectratio]{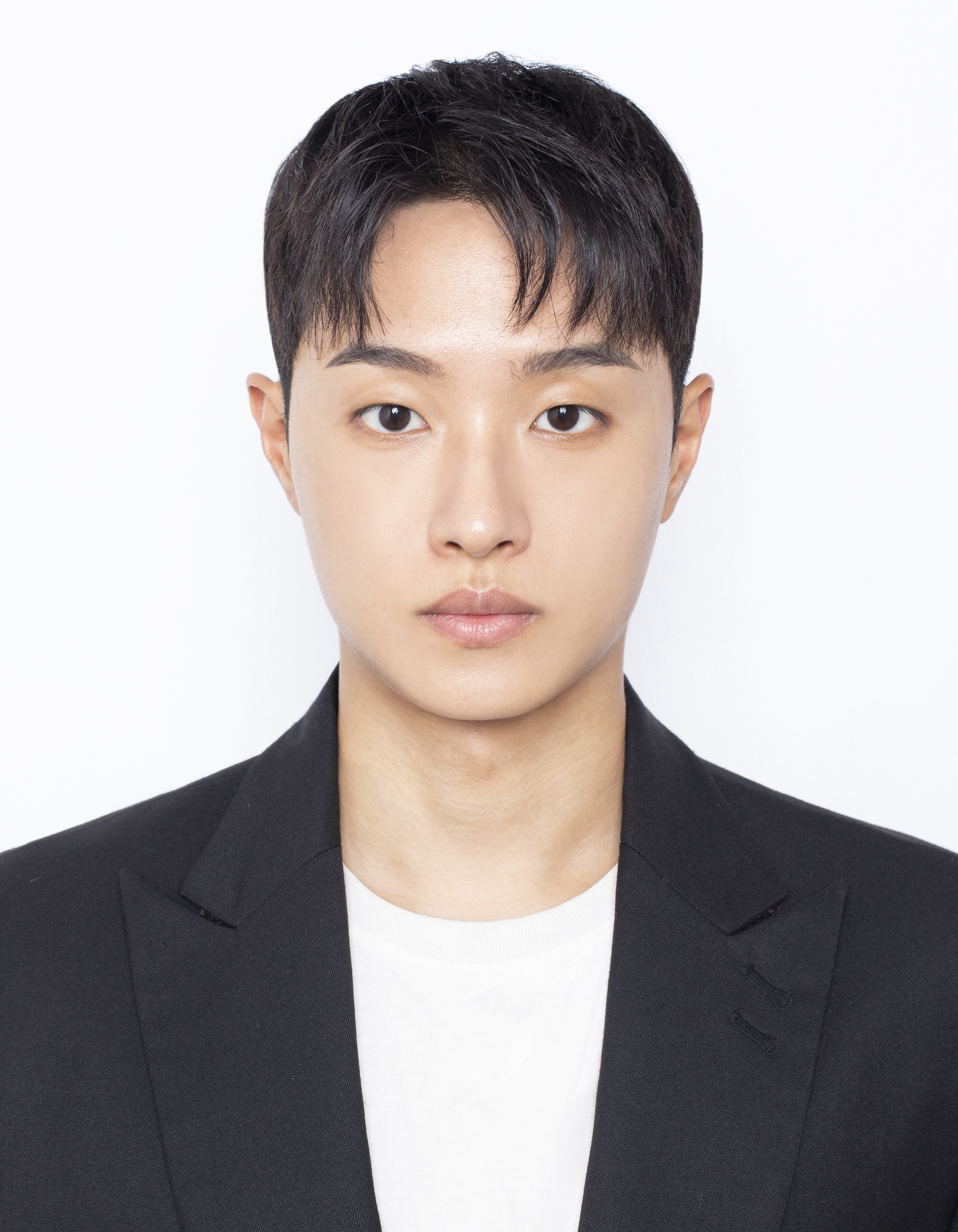}}]{Jihyong Oh} is an Assistant Professor in the Department of Imaging Science at the Graduate School of Advanced Imaging Science, Multimedia \& Film (GSAIM) at Chung-Ang University (CAU; Seoul, South Korea), and has been leading the Creative Vision and Multimedia Lab (https://cmlab.cau.ac.kr/) since September 2023. Prior to this, he was a postdoctoral researcher at VICLAB at KAIST (Daejeon, South Korea). He received his B.E., M.E., and Ph.D. degrees in Electrical Engineering from KAIST in 2017, 2019, and 2023, respectively. He was a research intern at Meta Reality Labs in 2022. His research primarily focuses on low-level vision, image/video restoration, 3D vision, and generative AI. He has published papers at CVPR, ICCV, ECCV, AAAI, TCSVT, and Remote Sensing, and serves as a reviewer for these conferences and journals, as well as for SIGGRAPH and IEEE TIP, TGRS, and Access. He received Outstanding Reviewer Awards for ICCV 2021 and CVPR 2024. 
\end{IEEEbiography} 

\vspace*{-5mm}
\begin{IEEEbiography}[{\includegraphics[width=1in,height=1.25in,clip,keepaspectratio]{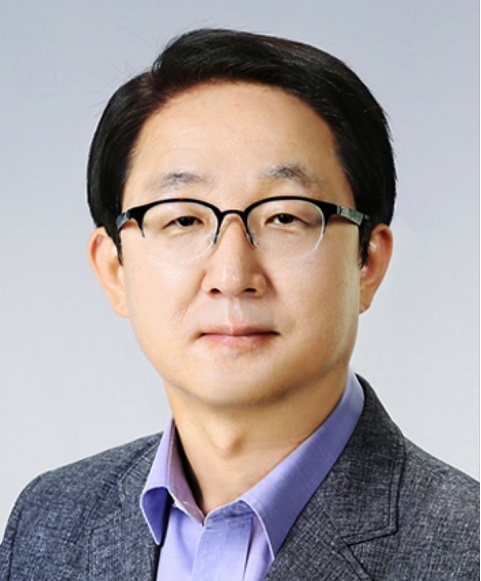}}]{Munchurl Kim}
(M’99–SM’13) received the B.E. degree in electronics from Kyungpook National University, Daegu, South Korea, in 1989, and the M.E. and Ph.D. degrees in electrical and computer engineering from the University of Florida, Gainesville, in 1992 and 1996, respectively. He joined the Electronics and Telecommunications Research Institute, Daejeon, South Korea, as a Senior Research Staff Member, where he led the Realistic Broadcasting Media Research Team. In 2001, he joined the School of Engineering, Information and Communications University (ICU), Daejeon, as an Assistant Professor. Since 2009, he has been with the School of Electrical Engineering, Korea Advanced Institute of Science and Technology (KAIST), Daejeon, where he is currently a Full Professor. He has published 212 (189) international (domestic) journal and conference papers. He holds 243 registered domestic and international patents in image restoration and video coding. He was invited to give a keynote speech on the evolution of conventional and deep video compression technologies in 2020 Multimedia Modeling Conference. He has served the technical program chairs of Visual Communications and Image Processing (VCIP) 2023 and ACM Multimedia Asia 2023. 
\end{IEEEbiography}

\vfill
\end{document}